\documentclass{article}
\usepackage{fancyhdr}

\usepackage{report}

\usepackage{soul}
\usepackage[utf8]{inputenc} 
\usepackage[T1]{fontenc}    
\usepackage{hyperref}       
\usepackage{url}            
\usepackage{booktabs}       
\usepackage{amsfonts}       
\usepackage{fancyhdr}       

\usepackage{nicefrac}       
\usepackage{microtype}      
\usepackage{graphicx}
\usepackage{pifont}
\usepackage{multirow}
\usepackage{CJKutf8}
\definecolor{darkmagenta}{rgb}{0.56, 0.0, 1.0}
\definecolor{softyellow}{rgb}{1.0, 0.92, 0.3} 
\definecolor{LightAquamarine}{rgb}{0.75, 1.0, 0.8} 
\definecolor{FireBrick}{RGB}{178,34,34}
\definecolor{MediumPurple}{RGB}{147,112,219}

\definecolor{uclablue}{rgb}{0.15, 0.45, 0.68}
\hypersetup{
    breaklinks,
    colorlinks=true,
    citecolor={darkmagenta},
    linkcolor={uclablue},
    urlcolor={uclablue}
}
\usepackage{wrapfig}
\usepackage{float}
\usepackage{subcaption}
\usepackage{placeins}
\usepackage{lipsum} 
\usepackage{tcolorbox}
\usepackage{amsmath}
\usepackage{amssymb}
\usepackage{utfsym}
\usepackage{fontawesome}
\usepackage{xspace}
\usepackage{enumitem}
\usepackage{multirow} 
\tcbuselibrary{breakable}
\usepackage{enumitem}
\usepackage{colortbl}
\usepackage{fancyhdr}

\usepackage{tcolorbox}
\usepackage{transparent}
\usepackage{wrapfig}

\newtcolorbox{promptbox}[1]{
  colback=gray!3,
  colframe=black!30,
  boxrule=0.5pt,
  arc=2pt,
  left=6pt,
  right=6pt,
  top=6pt,
  bottom=6pt,
  breakable,
  title={#1},
  colbacktitle=gray!15,
  coltitle=black,
  fonttitle=\bfseries,
}
\usepackage{xurl}
\usepackage{listings}
\lstdefinestyle{jsonstyle}{
  basicstyle=\ttfamily,
  breaklines=true,
  breakatwhitespace=false,
  columns=fullflexible,
  keepspaces=true,
  frame=single,
  showstringspaces=false
}

\definecolor{njuPurple}{RGB}{220,205,230}     
\definecolor{njuPurpleLight}{RGB}{250,245,252}   
\colorlet{njuPurpleMid}{njuPurpleLight!55!njuPurple}

\newtcolorbox{abstractbox}{
    colback=njuPurpleLight,   
    colframe=njuPurple,       
    boxrule=1pt,              
    arc=4mm,                  
    left=8pt,                 
    right=8pt,                
    top=8pt,                  
    bottom=8pt,               
    opacityback=0.95
}

\newcommand{\agent}[1]{\textsc{TVIR-Agent}}
\newcommand{\bench}[1]{\textsc{TVIR-Bench}}

\title{TVIR: Building Deep Research Agents Towards Text--Visual Interleaved Report Generation}

\author{
\textbf{Xinkai Ma$^{*}$}, \textbf{Zhiqi Bai$^{*}$}, \textbf{Dingling Zhang$^{*}$}, \textbf{Pei Liu}, \textbf{Yishuo Yuan}, \textbf{He Zhu}, \\ \textbf{Jiakai Wang}, \textbf{Qianqian Xie}, \textbf{Yifan Zhao}, \textbf{Xinlong Yang}, \textbf{Hao Cong}, \textbf{Zhiheng Yao}, \textbf{Fengxia Xie}, \textbf{Zihao Xu}, \textbf{Haoran Xu}, \textbf{Zhaohui Wang}, \textbf{Minghao Liu}, \textbf{Shirong Lin}, \textbf{Yingshui Tan}, \textbf{Yuchi Xu}, \textbf{Wenbo Su}, \textbf{Zhaoxiang Zhang}, \textbf{Bo Zheng}, \textbf{Jiaheng Liu$^{\dagger}$}\\[2mm]
Nanjing University \qquad Alibaba Group \\[2mm]
\texttt{maxinkai@smail.nju.edu.cn}\qquad\texttt{liujiaheng@nju.edu.cn}
}

\begin{document}

\maketitle
\let\oldthefootnote\thefootnote

\let\thefootnote\relax\footnotetext{*~Equal Contribution. ~~$^\dagger$~Corresponding Author.}
\let\thefootnote\oldthefootnote

\begin{abstractbox}
\begin{center}
\textbf{\Large Abstract}
\end{center}
Deep Research Agents have shown strong capability in multi-step information retrieval, reasoning, and long-form report generation, but existing benchmarks and systems remain predominantly text-centric, with limited evaluation of whether visual elements are factually reliable and well aligned with the surrounding analysis. To address this gap, we introduce \textbf{TVIR} (\textbf{T}ext--\textbf{V}isual \textbf{I}nterleaved \textbf{R}eport Generation), which includes \textbf{\bench{}}, a benchmark of 100 expert-curated multimodal deep research tasks that require visual elements to serve specific analytical sub-goals, and \textbf{\agent{}}, a hierarchical multi-agent framework that serves as a strong baseline for constructing outlines, retrieving images, generating charts with traceable sources, and composing reports through context-aware sequential writing. We further develop a dual-path evaluation framework that combines \textbf{Textual Assessment} and \textbf{Visual Assessment}. Experiments across nine deep research systems show that \agent{} achieves strong overall performance, underscoring the importance of explicit multimodal design and evaluation for evidence-driven report generation. The project page is available at \url{https://nju-link.github.io/TVIR}.

\end{abstractbox}

\section{Introduction}

Recent advances in large language models (LLMs) ~\citep{openai2024openaio1card,Guo_2025,kimiteam2026kimik2openagentic,5team2025glm45agenticreasoningcoding} have catalyzed the emergence of Deep Research Agents (DRAs)~\citep{zheng-etal-2025-deepresearcher,coelho2025deepresearchgymfreetransparentreproducible,cai2026yunquedeepresearchtechnicalreport,tongyideepresearchteam2025tongyideepresearchtechnicalreport}, which aim to autonomously conduct multi-step information retrieval, reasoning, and generation to produce comprehensive research reports. These systems have shown promising capability in long-context planning, citation grounding, and analytical writing, positioning them as potential assistants for complex professional decision-making in domains such as policy analysis, finance, and scientific research. As a result, a growing body of benchmarks and systems~\citep{shi2025deepresearchsystematicsurvey,zhang2025deepresearchsurveyautonomous} has focused on evaluating end-to-end deep research workflows rather than isolated questions.

\begin{wrapfigure}[31]{r}{0.58\textwidth}
    \centering
    \includegraphics[width=\linewidth]{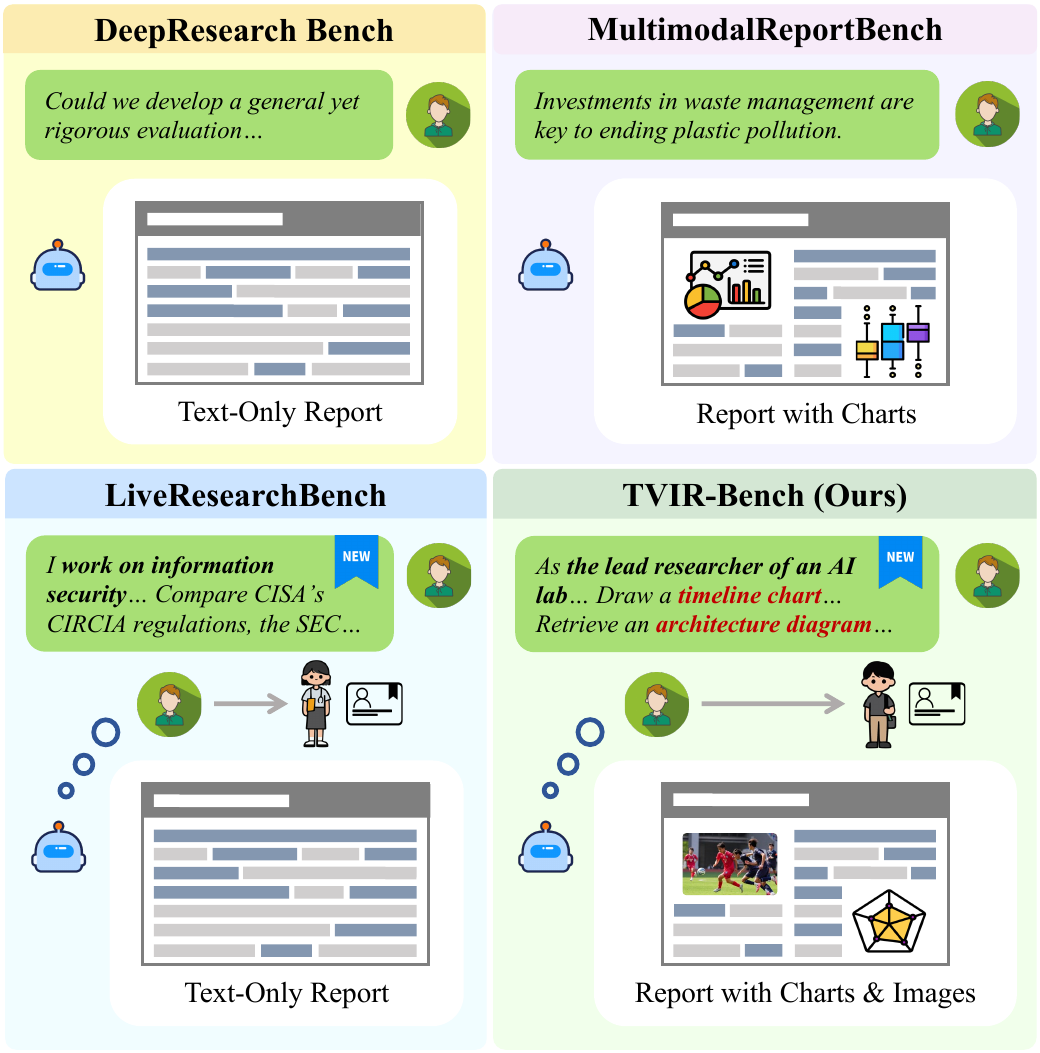}
    \caption{Comparison of representative deep research benchmarks. Existing benchmarks mainly focus on text-only or weakly multimodal reports, whereas \bench{} requires text--visual interleaved reports with semantically grounded charts and retrieved images.}
  \label{fig:intro}
\end{wrapfigure}

Despite this progress, as shown in Figure~\ref{fig:intro}, existing deep research paradigms remain predominantly text-centric. Most benchmarks and agent frameworks evaluate success based on textual coherence, depth, and citation support, while overlooking a critical characteristic of real-world professional reports: the integration of visual evidence. In practice, high-quality research outputs rarely rely on text alone. Instead, they interleave narrative analysis with charts, diagrams, and images that serve as evidential artifacts—supporting claims, revealing trends, and enabling rapid sense-making. When visual elements are present in current systems, they are often treated as decorative supplements rather than first-class reasoning components, with limited evaluation of their fidelity, provenance, or alignment with the surrounding text. This gap leads to
a fundamental mismatch between existing deep research benchmarks and the demands of real-world analytical work. A research agent that produces fluent text but inaccurate, misleading, or semantically disconnected visuals cannot be considered reliable for high-stakes decision-making. Moreover, without explicit evaluation protocols, models tend to optimize for superficial visual inclusion—prioritizing aesthetic completeness over evidential rigor. Addressing this limitation requires rethinking deep research not as a purely textual task, but as a multimodal synthesis problem in which text and visuals must be jointly generated, and evaluated.

{\parfillskip=0pt
To this end, we introduce \textbf{TVIR} (\textbf{T}ext--\textbf{V}isual \textbf{I}nterleaved \textbf{R}eport Generation), a unified benchmark and the corresponding agentic framework designed to advance deep research towards multimodal, evidence-driven report generation. First, we present \textbf{\bench{}}, a multimodal deep research benchmark
\par}

consisting of 100 expert-curated tasks spanning diverse domains and complexity levels. Unlike prior benchmarks, it enforces strict design principles that require visual elements—both retrieved images and code-generated charts—to be semantically grounded in specific analytical sub-goals, rather than appended post hoc.

Second, we propose \textbf{\agent{}}, a hierarchical multi-agent baseline framework for text–visual interleaved report generation. The agent decomposes user tasks into structured plans with explicit multimodal constraints, instantiates images and charts with traceable sources, and generates long-form reports through context-aware sequential writing. By explicitly modeling visual evidence throughout the planning and writing stages, \agent{} treats visuals as integral components of reasoning rather than optional embellishments.

Finally, we introduce a comprehensive evaluation suite that jointly audits textual and visual quality. Our framework combines \textbf{Textual Assessment} (focusing on citation grounding, logical consistency, and analytical depth) with \textbf{Visual Assessment}, which measures figure quality, chart fidelity, and cross-modal alignment between text and visuals. Through extensive experiments across multiple deep research systems, we highlight the insufficiency of existing paradigms and underscore the need for multimodal deep research agent designs.

In summary, our contributions are threefold :
\begin{itemize}
    \item We introduce \bench{}, the first comprehensive benchmark specifically designed to evaluate the end-to-end generation of long-context multimodal research reports, which establishes a new foundation for developing and evaluating deep research agents.
    \item 
    We propose \agent{}, an autonomous framework designed to handle the text--visual interleaved report generation. The system consists of a Planner, a Visual Asset Instantiation module, a Writer and a Polisher. 
    \item 
    We develop a rigorous dual-path evaluation framework that assesses reports through Textual Assessment and Visual Assessment. Our extensive experiments with nine representative deep research systems provide the critical insight that while current LLMs excel at textual fluency, they frequently prioritize "decorative" visuals over "evidential" ones, highlighting a significant gap in existing paradigms for evidence-based multimodal reasoning.
\end{itemize}

\section{Related Work}

\paragraph{\textbf{Deep Research Agent}}

Deep Research Agents (DRAs) have become an important paradigm for long-horizon retrieval, reasoning, and report generation. Existing DRAs, such as WebThinker~\citep{NEURIPS2025_ae03bdef} and WebWeaver~\citep{li2025webweaverstructuringwebscaleevidence}, are mostly text-centric. Although some recent work explores multimodal settings, such as Multimodal DeepResearcher~\citep{Yang_Pan_Wang_Wang_Liu_Weng_Feng_Feng_Zhu_Zhang_2026} and FinSight~\citep{jin2025finsightrealworldfinancialdeep}, visuals are still typically treated as auxiliary outputs and mainly rely on charts. In contrast, \agent{} formulates deep research as text--visual interleaved report generation, integrating retrieved images and code-generated charts into the full research workflow.

\paragraph{\textbf{Deep Research Benchmark}}

Existing deep research benchmarks mainly evaluate text-only report generation, such as DeepResearch Bench~\citep{du2025deepresearchbenchcomprehensivebenchmark} and DeepResearch Bench \MakeUppercase{\romannumeral 2}~\citep{li2026deepresearchbenchiidiagnosing}. In multimodal settings, MultimodalReportBench~\citep{Yang_Pan_Wang_Wang_Liu_Weng_Feng_Feng_Zhu_Zhang_2026} considers reports with code-generated charts, but its evaluation remains coarse-grained. MMDeepResearch-Bench~\citep{huang2026mmdeepresearchbenchbenchmarkmultimodaldeep} further studies image--text inputs, where report visuals are grounded in task-provided images. In contrast, \bench{} evaluates end-to-end text-visual interleaved report generation with fine-grained textual and visual assessment.

\section{\bench{}}

\subsection{Data Design}
\label{sec:data_design}

\begin{wrapfigure}{r}{0.5\textwidth}
\centering
\includegraphics[width=\linewidth]{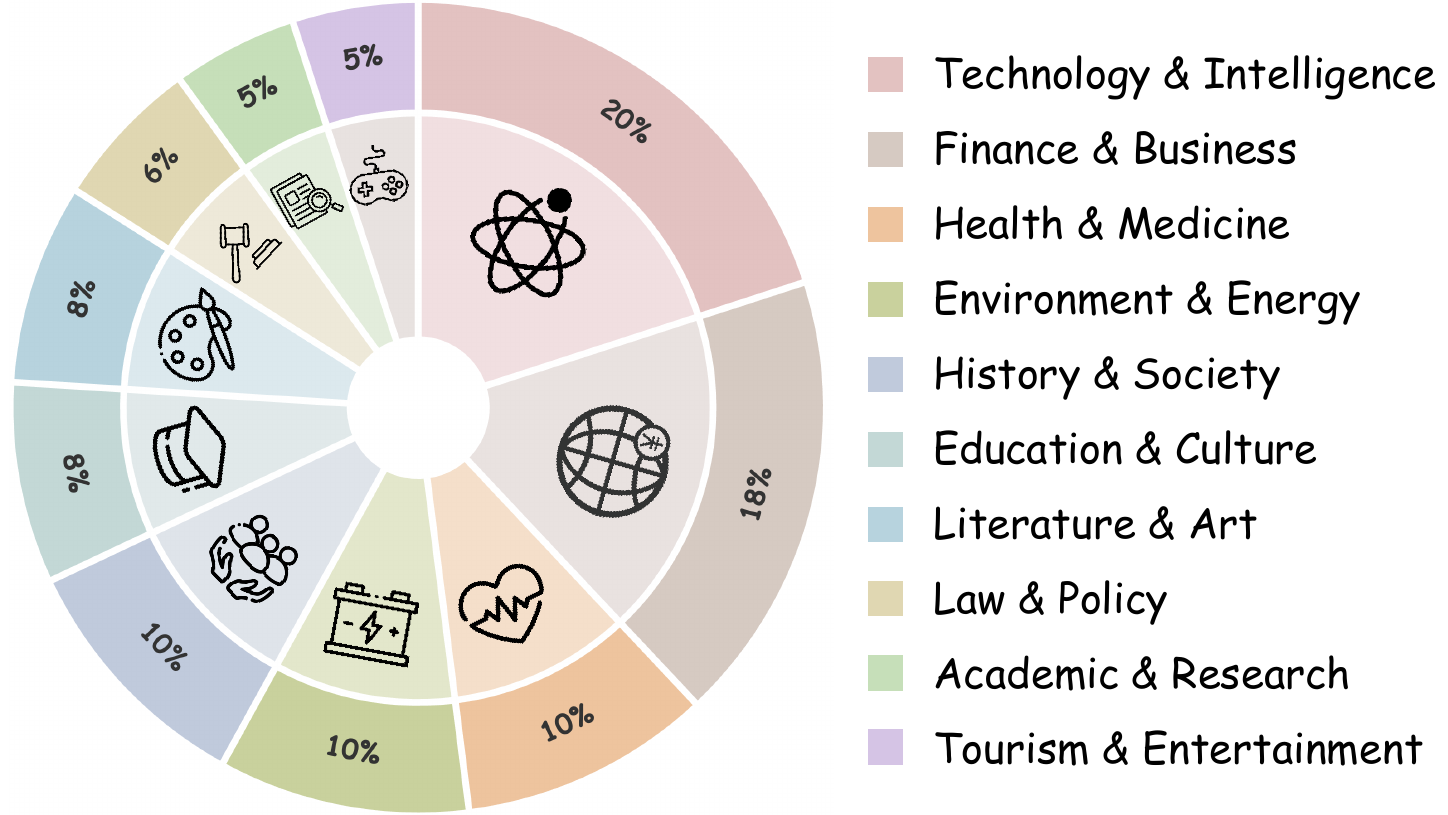}
\caption{Domain taxonomy of \bench{}.}

\label{fig:domain}
\end{wrapfigure}

\paragraph{Domain Coverage}To ensure comprehensive and representative domain coverage, we draw on the domain taxonomies and task distributions of several existing deep research benchmarks, including DeepResearch Bench~\citep{du2025deepresearchbenchcomprehensivebenchmark}, LiveResearchBench~\citep{wang2025liveresearchbenchlivebenchmarkusercentric}, and DeepResearchEval~\citep{wang2026deepresearchevalautomatedframeworkdeep}, and develop the domain taxonomy shown in Figure~\ref{fig:domain}. While preserving coverage of the humanities and basic sciences, we place greater emphasis on \textit{Technology \& Intelligence} and \textit{Finance \& Business} to better align the needs of high-stakes decision-making.

\paragraph{Task Design and Complexity}Task construction is guided by five core design principles: \textbf{role-driven}, \textbf{demand-oriented}, \textbf{deep research}, \textbf{frontier-focused}, and \textbf{multimodal integration}. These principles ensure that tasks are grounded in realistic user needs, require substantive analytical synthesis rather than simple information retrieval, and incorporate explicit multimodal elements to better reflect real-world deep research workflows. To systematically evaluate model performance under varying cognitive demands, we further organize tasks into three complexity levels, corresponding to low, medium, and high requirements for multimodal integration and instruction following. Further details are provided in Appendix~\ref{app:a}.

\subsection{Data Construction}
\label{sec:data_construction}

\begin{figure*}[t]
\centering
\includegraphics[width=\linewidth]{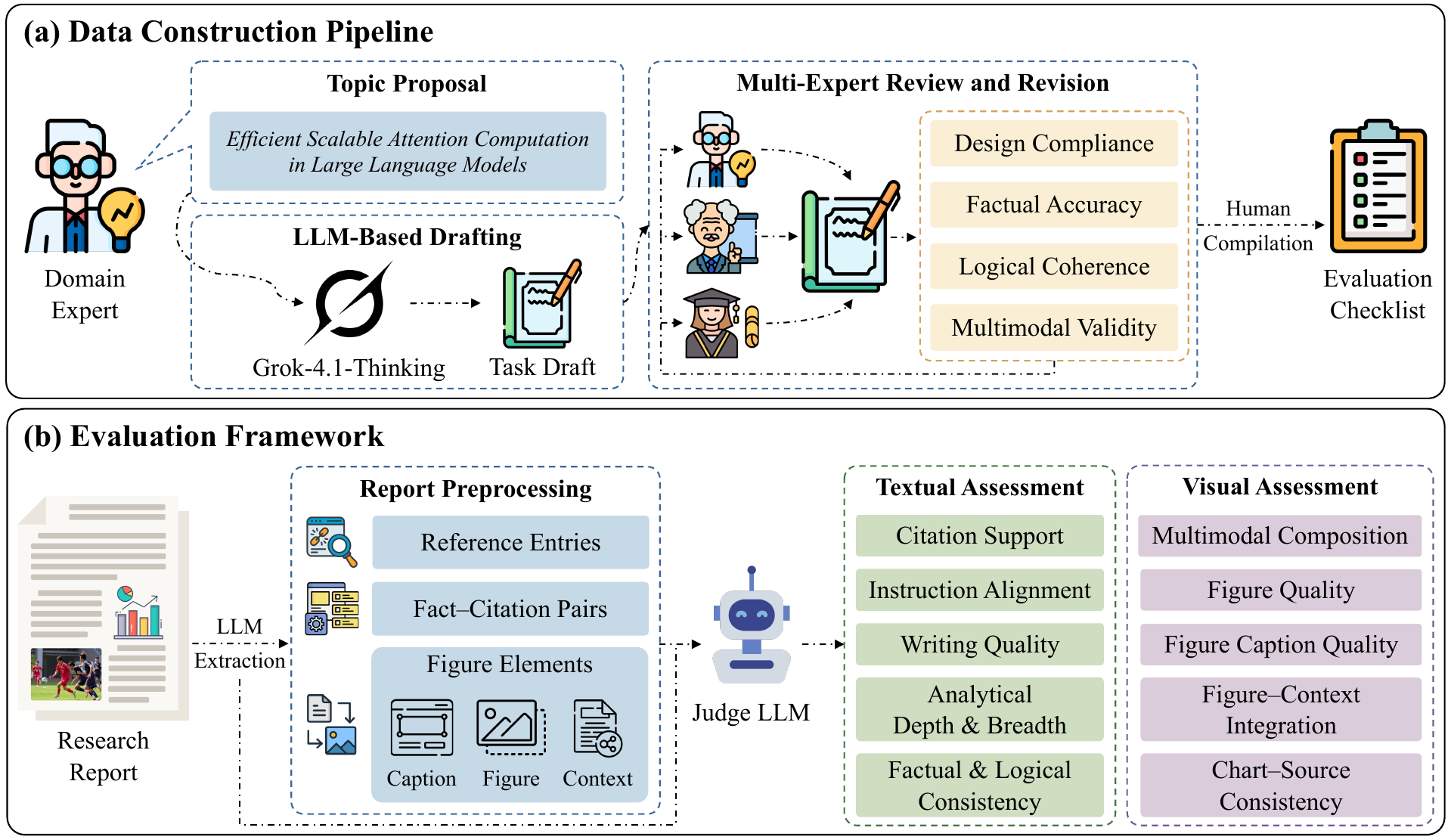}
\caption{Overview of \bench{}, including data construction pipeline and evaluation framework.}
\label{fig:pipeline}
\end{figure*}

As shown in Figure~\ref{fig:pipeline}(a), dataset construction follows an expert-driven workflow with four stages.

\paragraph{Expert Topic Proposal}
For each task, a domain expert with sufficient professional knowledge and judgment proposes a core topic. To satisfy the frontier-focused principle defined above, the topic must be reasonably novel and timely.

\paragraph{LLM-Based Task Drafting}
Based on the core topic proposed by the domain expert, we use Grok- 4.1-Thinking to draft the task. We guide the model to produce tasks that are logically coherent, researchable, and practically meaningful. The detailed prompt is provided in Appendix~\ref{app:task_drafting_prompt}.

\paragraph{Multi-Expert Review and Revision}
Each draft is then examined by three domain experts and revised accordingly. The review focuses on four aspects: \textbf{(1) Design Compliance}, which checks whether the task satisfies the predefined task design principles; \textbf{(2) Factual Accuracy}, which verifies the correctness of entities, concepts, claims, and contextual details involved in the task; \textbf{(3) Logical Coherence}, which evaluates whether the task is clearly formulated and whether its sub-questions form a meaningful and well-connected whole; and \textbf{(4) Multimodal Validity}, which ensures that required multimodal elements are practically obtainable, including publicly retrievable images and code-generated charts based on real and accessible data.

\paragraph{Checklist Compilation}
For each accepted task, we compile a corresponding evaluation checklist. The checklist converts the task into a set of verifiable items for systematically assessing whether a generated report fully and accurately addresses its requirements. Each item is formulated as a small number of clear and actionable atomic checkpoints, without going beyond the task itself.

\subsection{\textbf{Dataset Statistics}}

\bench{} comprises 100 high-quality multimodal deep research tasks, including 50 Chinese tasks and 50 English tasks. They span 10 major domains and are proportionally balanced across the three predefined complexity levels. The sub-questions within these tasks cover eight high-level functional types, such as trend prediction, mechanism explanation, and comparative analysis, with a roughly balanced distribution across the dataset.

\subsection{Evaluation Framework}

As shown in Figure~\ref{fig:pipeline}(b), we propose a multi-dimensional evaluation framework for end-to-end auditing of generated research reports. It consists of two complementary components: \textbf{Textual Assessment (TA)} and \textbf{Visual Assessment (VA)}. TA and VA are computed as the arithmetic mean of their respective fine-grained metric scores. Unless otherwise specified, all metrics are evaluated using an LLM-as-a-Judge, with scores normalized to a scale of 0--100. See Appendix~\ref{app:fine_grained_metrics} for metric computation details and Appendix~\ref{app:judge_prompts} for evaluation prompts.

\subsubsection{Report Preprocessing}

To support downstream evaluation, we preprocess each report to extract structured information using the judge LLM. Specifically, we extract (i) \textit{reference entries}, including their reference indices and associated URLs, (ii) \textit{fact--citation pairs} for textual assessment, where factual statements are linked to citation indices, and (iii) \textit{figure elements} for visual assessment, together with their captions, base64-encoded visual content, surrounding context, and associated citation indices. We then use the extracted URLs to retrieve the corresponding webpage text via the Serper API. The extraction prompts are provided in Appendix~\ref{app:preprocessing_prompts}.

\subsubsection{\textbf{Textual Assessment (TA)}}

\textbf{Citation Support (CS)} measures whether factual statements are supported by their cited references. Using the extracted fact--citation pairs and retrieved reference contents, the judge LLM assigns each fact a score of $1$, $0.5$, or $0$ for \textit{supported}, \textit{partially supported}, and \textit{unsupported}, respectively. 

\textbf{Instruction Alignment (IA)} evaluates whether the report satisfies the task checklist. For each checklist item, the judge LLM assigns a score of $1$, $0.5$, or $0$ based on how completely and specifically the report addresses it.

\textbf{Writing Quality (WQ)}, \textbf{Analytical Depth \& Breadth (ADB)}, and \textbf{Factual \& Logical Consistency (FLC)} are assessed at the report level. WQ evaluates coherence and organization, clarity and readability, conciseness, and stylistic consistency. ADB assesses whether the report goes beyond surface-level description through explanatory reasoning, sustained analysis, critical evaluation, forward-looking insight, and broad thematic coverage. FLC measures self-consistency by detecting factual or logical contradictions and mapping the resulting issue count to a discrete score. 

\subsubsection{\textbf{Visual Assessment (VA)}}

\textbf{Multimodal Composition (MC)} provides a report-level assessment of how effectively figure elements are organized across the report, considering their layout, quantity, variety, and richness.

\textbf{Figure Quality (FQ)} captures the intrinsic visual quality of figures. It combines \textbf{Image Quality}, derived from CV-based measurements of resolution, aspect ratio, sharpness, and contrast together with a duplication penalty, and \textbf{Chart Quality}, assessed by the judge LLM using binary checklist evaluations of layout integrity, readability, and conciseness. The final FQ score is computed as a count-weighted average of the two component scores.

\textbf{Figure Caption Quality (FCQ)}, \textbf{Figure--Context Integration (FCI)}, and \textbf{Chart--Source Consistency (CSC)} are assessed for each extracted figure element. FCQ evaluates whether a caption accurately describes the figure, provides sufficient interpretive information, and remains clear and readable. FCI assesses how well each figure relates to its surrounding text, is incorporated into the narrative flow, and contributes information beyond what text alone can effectively convey. CSC measures the consistency of each chart with its cited sources by identifying contradictions and mapping the issue count to a discrete score.

\begin{figure}[t]
\centering
\includegraphics[width=\linewidth]{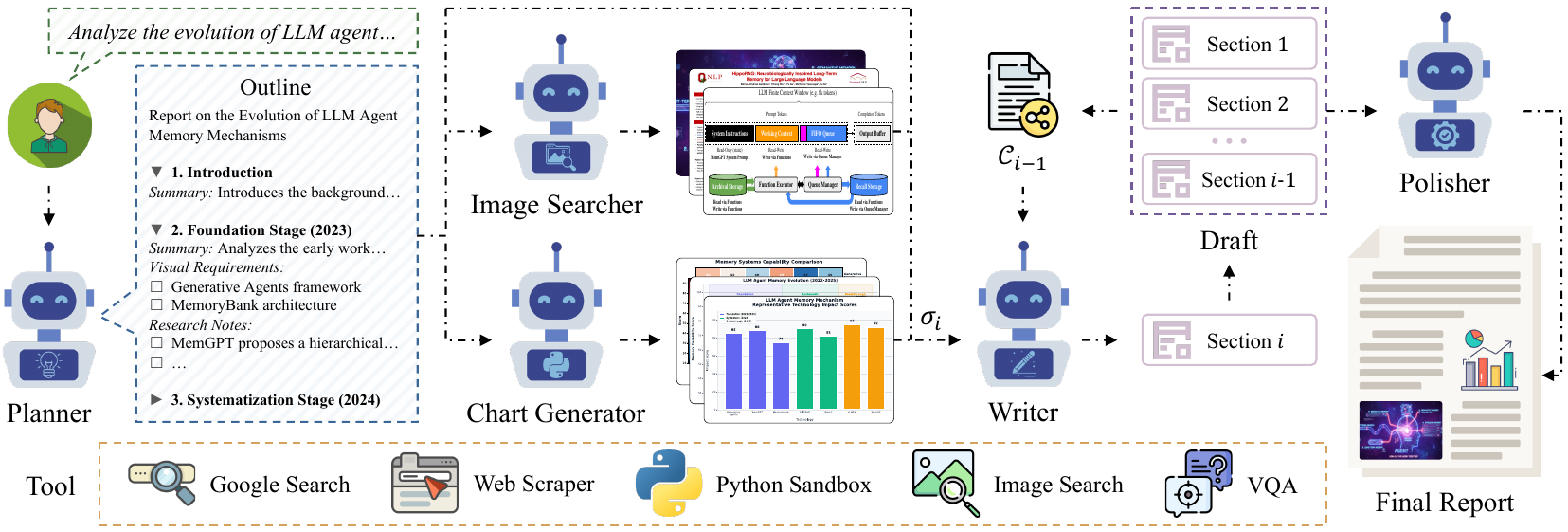}
\caption{Overall architecture of \agent{} for text--visual interleaved report generation.}
\label{fig:agent}
\end{figure}

\section{\agent{}}
\label{section:TVIR_agent}

We develop \agent{} based on MiroThinker ~\citep{miromindteam2025mirothinkerpushingperformanceboundaries} for text--visual interleaved report generation. Figure~\ref{fig:agent} illustrates its overall architecture. Given a user task $\mathcal{T}$, \agent{} produces a high-quality research report $\mathcal{R}$ through four stages, which we introduce below.

\subsection{\textbf{Research-Grounded Planning}}

As the initialization module, the \textbf{Planner} parses the user task $\mathcal{T}$ and iteratively invokes external tools such as Google Search and web scraping to retrieve relevant information. It then synthesizes the collected information into a structured outline $\mathcal{O} = \{\sigma_1, \sigma_2, \dots, \sigma_N\}$, where each outline unit $\sigma_i$ contains a section title, a brief summary, planned visual requirements $\mathcal{V}_i^{\mathrm{req}}$, and section-level research notes $\mathcal{N}_i$. Each note $n \in \mathcal{N}_i$ records a citation, source URL, and key findings. These notes provide factual grounding for subsequent stages and improve both credibility and traceability.

\subsection{\textbf{Visual Asset Instantiation}}

Given the planned outline $\mathcal{O}$, this stage instantiates the visual requirements of each section. To address different visual needs, we employ two specialized agents. The \textbf{Image Searcher} handles visual concepts such as portraits, scenes, and architecture diagrams by retrieving candidate images through Google Image Search, filtering low-quality results with heuristic rules, and using a visual question answering (VQA) tool for relevance verification before selecting the most suitable one. The \textbf{Chart Generator} handles content involving data distributions or relationships by retrieving relevant data through search and web scraping tools, verifying authenticity and cross-source consistency, generating Python plotting code, and executing it in a sandbox to produce charts.

After this stage, each outline unit $\sigma_i$ is updated with instantiated visual assets $\mathcal{V}_i^{\mathrm{inst}}$, together with captions, descriptions, and source provenance, yielding the augmented outline $\mathcal{O}^{\mathrm{vis}}$. For code-generated charts, the original data-source URLs are preserved; for retrieved images, the corresponding source webpage URLs are retained.

\subsection{\textbf{Context-Aware Sequential Writing}}

The \textbf{Writer} generates the report section by section. To maintain coherence across sections and reduce redundancy, it conditions on the current outline unit $\sigma_i$ and a dynamically updated global context $\mathcal{C}_{i-1}$, which consists of the titles, summaries, and subsection structures of previously generated sections. It also uses the associated research notes $\mathcal{N}_i$ as supporting evidence and, when they are insufficient, invokes search and web scraping tools to gather additional verifiable information.

During generation, the Writer determines insertion points for the instantiated visual assets in $\sigma_i$ based on their descriptions and composes Markdown content with interleaved text and visual elements, using local paths for code-generated charts and online URLs for retrieved images. Each section maintains its own figure and citation numbering, both starting from 1.

\subsection{\textbf{Global Index Polishing}}

In the final stage, the \textbf{Polisher} processes references and figures at the report level. It first removes uncited references from each section, then deduplicates the remaining references globally by URL and normalized content, and renumbers them into a unified reference list, while updating in-text citation markers accordingly. Finally, it renumbers figures across sections, reassigns figure IDs and labels in sequential order, and updates in-text figure references to match.

\section{Experiments}
\subsection{Experimental Setup}

We evaluate a total of nine deep research systems, including six commercial closed-source systems and three \agent{} variants built with different backbone LLMs. The commercial systems consist of one text-only report generation system, Gemini-3-Pro Deep Research~\citep{gemini}, and five text--visual interleaved report generation systems: Grok-4.1-Thinking DeepSearch~\citep{grok}, Claude-4.5-Sonnet w/Search~\citep{claude}, Perplexity Deep Research~\citep{perplexity}, Genspark Deep Research~\citep{genspark}, and Manus-1.6~\citep{manus}. The three \agent{} variants use Qwen3-Max~\citep{qwen}, GLM-4.7~\citep{glm}, and Claude-4.5-Sonnet as backbone LLMs. More details on report collection are provided in Appendix~\ref{app:report_collection}. Report preprocessing and all LLM-based evaluation are conducted using GPT-5.2~\citep{gpt} as the judge LLM, with temperature set to 0.

\subsection{Main Results}

\paragraph{Overall performance}
Table~\ref{tab:main} presents the main results on \bench{}. Overall, \agent{} variants achieve the strongest aggregate performance among all evaluated systems. In particular, TVIR-Agent (Claude-4.5-Sonnet) obtains the best Overall score, followed by TVIR-Agent (Qwen3-Max) and TVIR-Agent (GLM-4.7), while Manus-1.6 is the strongest commercial system. The aggregate metrics also reveal different system strengths: TVIR-Agent (GLM-4.7) achieves the best TA score, showing strong textual capabilities, whereas TVIR-Agent (Claude-4.5-Sonnet) achieves the best VA score by a clear margin, demonstrating superior visual grounding and cross-modal alignment. These results indicate that our framework is particularly effective for deep research tasks requiring both textual synthesis and visual integration.

\paragraph{Key insights}
The results also reveals several insights into current deep research systems. First, no single model consistently performs best on every fine-grained dimension, suggesting that strong aggregate performance does not necessarily translate into consistent strength across all aspects of text--visual research. Second, several commercial systems remain competitive on textual assessment metrics, indicating that high-quality textual synthesis is already a relative strength of existing deep research products. Finally, Gemini-3-Pro Deep Research cannot be evaluated on VA and Overall because it generates text-only reports, which further highlights the importance of native multimodal support in \agent{}.

\begin{table*}[t]
  \centering
  \resizebox{\textwidth}{!}{
  \begin{tabular}{l ccccc ccccc ccc}
    \toprule
    \multirow{2}{*}{\textbf{System}} 
    & \multicolumn{5}{c}{\textbf{Textual Assessment}} 
    & \multicolumn{5}{c}{\textbf{Visual Assessment}} 
    & \multicolumn{3}{c}{\textbf{Aggregate}} \\
    \cmidrule(lr){2-6} \cmidrule(lr){7-11} \cmidrule(lr){12-14}
    \cmidrule(lr){2-6} \cmidrule(lr){7-11} \cmidrule(lr){12-14}
    & CS & IA & WQ & ADB & FLC
    & FQ & MC & FCQ & FCI & CSC
    & \textbf{TA} & \textbf{VA} & \textbf{Overall} \\
    \midrule
    \multicolumn{14}{c}{\textbf{Commercial Closed-Source Systems}}\\
    \midrule
    Gemini-3-Pro Deep Research & 14.96 & 58.31 & 66.88 & 63.94 & 88.50 & - & - & - & - & - & 58.52 & - & -\\
    Grok-4.1-Thinking DeepSearch & 17.72 & 60.65 & 67.68 & 57.58 & \underline{89.20} & 80.43 & 52.15 & 47.04 & 46.76 & 5.75 & 58.56 & 46.43 & 52.49\\
    Claude-4.5-Sonnet w/Search & 47.53 & 79.32 & 69.37 & 70.52 & 84.00 & 90.24 & 63.85 & 61.47 & 53.43 & 67.49 & \underline{70.15} & 67.30 & 68.72\\
    Perplexity Deep Research & 44.60 & 81.03 & 67.60 & \underline{70.64} & 80.90 & 73.62 & 62.90 & 59.02 & 63.41 & 8.35 & 68.95 & 53.46 & 61.20 \\
    Genspark Deep Research & 35.27 & \textbf{83.71} & 69.28 & \underline{70.64} & 84.60 & \textbf{92.87} & 70.70 & 63.58 & 59.00 & 40.28 & 68.70 & 65.29 & 66.99\\
    Manus-1.6 & 45.57 & 74.12 & \textbf{72.15} & 62.84 & \textbf{92.40} & 86.27 & \underline{70.75} & 66.14 & 71.03 & 56.02 & 69.42 & 70.04 & 69.73\\
    \midrule
    \multicolumn{14}{c}{\textbf{\agent{} Variants (Ours)}}\\
    \midrule
    TVIR-Agent (Qwen3-Max) & \underline{53.68} & 76.69 & 69.30 & 67.48 & 83.00 & \underline{91.71} & 67.80 & \underline{72.44} & \underline{74.56} & \textbf{78.63} & 70.03 & \underline{77.03} & \underline{73.53}\\
    TVIR-Agent (GLM-4.7) & \textbf{68.64} & 71.98 & 69.20 & 68.16 & 80.20 & 84.61 & 62.55 & 70.13 & 73.39 & 77.35 & \textbf{71.64} & 73.61 & 72.62\\
    TVIR-Agent (Claude-4.5-Sonnet) & 51.20 & \underline{81.09} & \underline{69.88} & \textbf{72.22} & 76.20 & 87.17 & \textbf{77.80} & \textbf{74.49} & \textbf{76.75} & \underline{77.58} & 70.12 & \textbf{78.76} & \textbf{74.44}\\
    
    \bottomrule
    \end{tabular}
    }
  \caption{Main results on \bench{}, averaged over three independent runs. \textbf{Bold} and \underline{underlined} indicate the best and second-best performance on each dimension, respectively. Overall is computed as the mean of TA and VA.}
  \label{tab:main}
\end{table*}

\subsection{Further Analysis}

\begin{wrapfigure}[35]{r}{0.56\textwidth}
  \includegraphics[width=\linewidth]{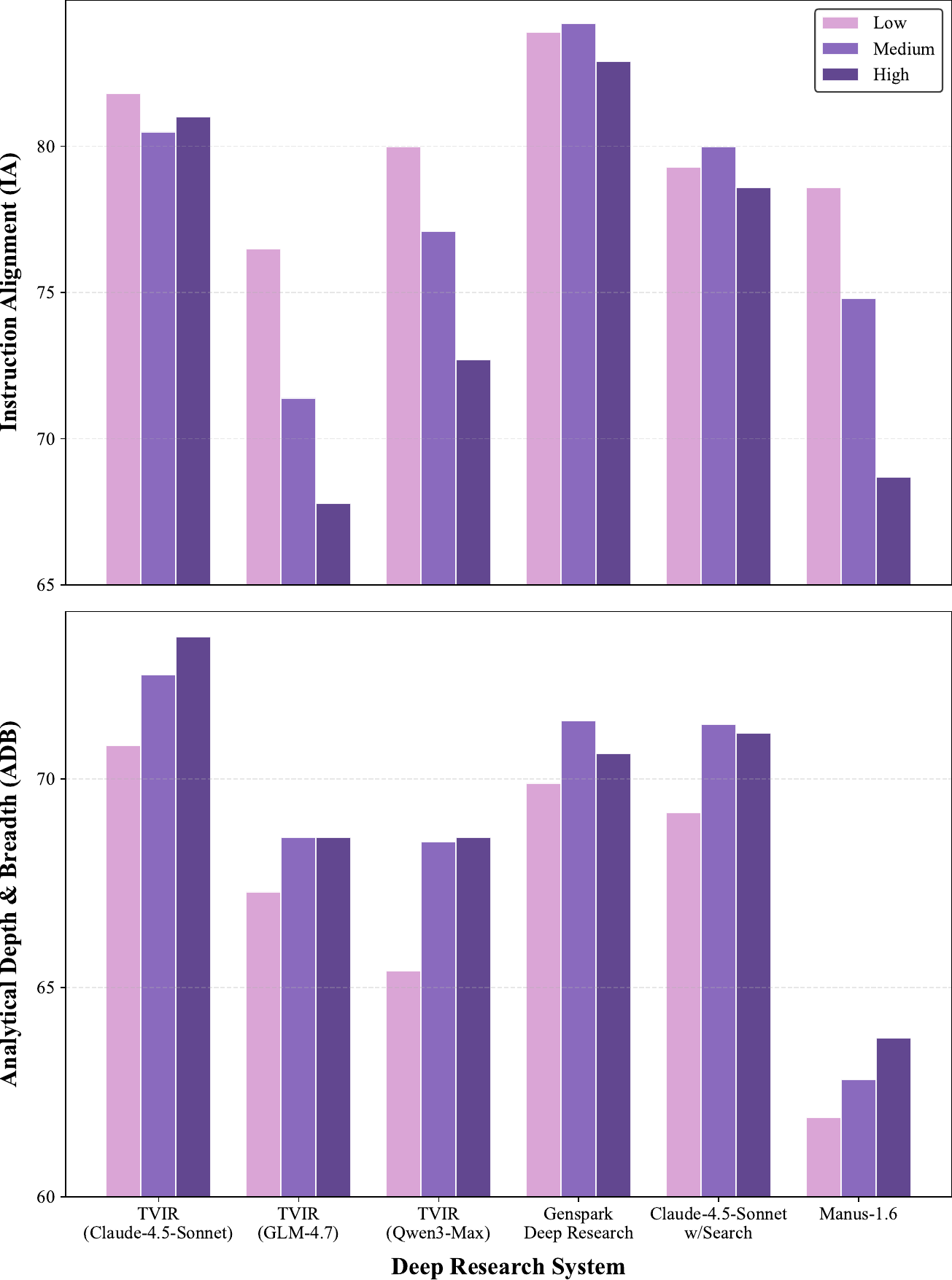}
  \caption{IA and ADB performance across task complexity levels on \bench{}.}
  \label{fig:level}
\end{wrapfigure}

\paragraph{Fine-Grained Strengths and Limitations}
Compared with commercial systems, \agent{} variants show clearer advantages in CS and most VA dimensions. On \textbf{CS}, TVIR-Agent (GLM-4.7) achieves 68.64, outperforming the best commercial system Claude-4.5-Sonnet w/Search (47.53) by \textbf{21.11} points. For \textbf{FCQ}, TVIR-Agent (Claude-4.5-Sonnet) scores 74.49, exceeding Manus-1.6 (66.14) by \textbf{8.35} points. These results indicate better evidence grounding and more reliable cross-modal alignment in \agent{}. By contrast, \textbf{FLC} remains relatively weaker for several strong systems, likely because longer and more detailed reports are inherently harder to maintain consistently. This suggests that long-form factual and logical consistency is still a shared challenge.

\paragraph{Performance across Task Complexity Levels}
Figure~\ref{fig:level} shows a consistent trend across \agent{} variants and leading commercial systems. As task complexity increases, \textbf{IA} generally declines, while \textbf{ADB} tends to improve. This pattern suggests that more complex tasks place heavier demands on multimodal coordination, fine-grained instruction tracking, and long-horizon reasoning, making it more difficult for research systems to fully satisfy task requirements. At the same time, such tasks appear to encourage more comprehensive and exploratory responses, leading to greater explanatory depth, broader thematic coverage, and more sustained analysis.

\begin{table*}[t]
\centering
\small
\begin{tabular}{lcccccc}
\toprule
\textbf{System} & \textbf{TA (ZH)} & \textbf{TA (EN)} & \textbf{VA (ZH)} & \textbf{VA (EN)} & \textbf{Overall (ZH)} & \textbf{Overall (EN)} \\
\midrule
\multicolumn{7}{c}{\textbf{Commercial Closed-Source Systems}}\\
\midrule
Gemini-3-Pro Deep Research     & 59.64 & 57.39 & --    & --    & --    & --    \\
Grok-4.1-Thinking DeepSearch   & 58.84 & 58.29 & 38.65 & 54.20 & 48.75 & 56.24 \\
Claude-4.5-Sonnet w/Search     & 70.84 & 69.46 & 69.61 & 64.98 & 70.23 & 67.22 \\
Perplexity Deep Research       & 69.13 & 68.78 & 52.90 & 54.02 & 61.01 & 61.40 \\
Genspark Deep Research         & 69.29 & 68.11 & 64.20 & 66.37 & 66.75 & 67.24 \\
Manus-1.6                      & \underline{71.10} & 67.74 & \underline{71.82} & 68.27 & \underline{71.46} & 68.00 \\
\midrule
\multicolumn{7}{c}{\textbf{\agent{} Variants (Ours)}}\\
\midrule
TVIR-Agent (Qwen3-Max)         & 71.01 & 69.04 & 77.24 & \underline{76.82} & 74.13 & \underline{72.93} \\
TVIR-Agent (GLM-4.7)           & \textbf{72.91} & \textbf{70.36} & 74.41 & 72.81 & 73.66 & 71.59 \\
TVIR-Agent (Claude-4.5-Sonnet) & 70.60 & \underline{69.63} & \textbf{79.64} & \textbf{77.88} & \textbf{75.12} & \textbf{73.76} \\
\bottomrule
\end{tabular}
\caption{Results across languages on \bench{}.}
\label{tab:lang}
\end{table*}

\begin{table*}[!htbp]
  \centering
  \resizebox{\textwidth}{!}{
  \begin{tabular}{l cccccccccc}
    \toprule
    \textbf{System} & \textbf{Tech. \& Intell.} & \textbf{Fin. \& Bus.} & \textbf{Health \& Med.} & \textbf{Law \& Policy} & \textbf{Env. \& Energy} & \textbf{Acad. \& Res.} & \textbf{Edu. \& Cult.} & \textbf{Lit. \& Art} & \textbf{Hist. \& Soc.} & \textbf{Tour. \& Ent.} \\
    \midrule
    \multicolumn{11}{c}{\textbf{Commercial Closed-Source Systems}}\\
    \midrule
    Grok-4.1-Thinking DeepSearch & 51.40 & 52.78 & 53.93 & 54.68 & 52.55 & 56.44 & 50.86 & 53.37 & 55.64 & 41.26\\
    Claude-4.5-Sonnet w/Search & 67.09 & 68.87 & 69.91 & 69.36 & 67.87 & 65.93 & 69.59 & 69.32 & 70.75 & 69.66\\
    Perplexity Deep Research & 62.07 & 59.95 & 59.03 & 61.80 & 61.21 & 61.58 & 62.45 & 60.41 & 62.99 & 61.23\\
    Genspark Deep Research & 66.21 & 66.39 & 66.97 & 65.88 & 68.83 & 67.46 & 68.35 & 65.51 & 67.34 & 69.05\\
    Manus-1.6 & 68.48 & 70.72 & 68.51 & 65.68 & 72.06 & \underline{73.58} & 63.13 & 70.60 & \underline{74.10} & 70.39\\
    \midrule
    \multicolumn{11}{c}{\textbf{\agent{} Variants (Ours)}}\\
    \midrule
    TVIR-Agent (Qwen3-Max) & \textbf{73.38} & \underline{71.70} & \underline{73.55} & \textbf{74.34} & \underline{73.83} & 71.57 & \textbf{75.63} & \underline{75.09} & 73.52 & \textbf{75.20}\\
    TVIR-Agent (GLM-4.7) & \textbf{73.44} & 70.86 & \textbf{74.28} & 72.11 & \textbf{74.82} & 67.66 & 74.33 & \textbf{75.27} & 69.42 & 72.96\\
    TVIR-Agent (Claude-4.5-Sonnet) & \underline{73.16} & \textbf{75.27} & 73.73 & \underline{73.45} & 74.34 & \textbf{75.40} & \textbf{75.63} & 74.95 & \textbf{74.90} & \underline{74.70}\\
    \bottomrule
    \end{tabular}
  }
  \caption{Overall results across domains on \bench{}.}
  \label{tab:domain}
\end{table*}

\paragraph{Performance across Languages}
Table~\ref{tab:lang} shows a mild overall advantage on the Chinese subset, especially in textual assessment, where all evaluated systems achieve higher TA scores than on English tasks. For VA and Overall, most systems also perform slightly better on Chinese, though a few, including Grok-4.1-Thinking DeepSearch, Perplexity Deep Research, and Genspark Deep Research, achieve comparable or stronger results on English. The gap is generally modest, and system rankings remain broadly stable across languages. One possible reason is that the two subsets are not direct translations, but are designed with language-specific cultural and real-world context. They should therefore be viewed as parallel benchmark slices rather than strictly matched pairs. Despite these differences, \agent{} variants consistently rank among the top systems on both subsets, suggesting strong cross-lingual generalization.

\begin{wraptable}{r}{0.4\textwidth}
\centering
\begin{tabular}{lc}
\toprule
\textbf{Domain} & \textbf{Avg. Rank} \\
\midrule
History \& Society         & 3.88 \\
Education \& Culture       & 4.12 \\
Tourism \& Entertainment   & 5.00 \\
Environment \& Energy      & 5.12 \\
Literature \& Art          & 5.25 \\
Academic \& Research       & 5.50 \\
Health \& Medicine         & 5.88 \\
Law \& Policy              & 6.25 \\
Finance \& Business        & 6.62 \\
Technology \& Intelligence & 7.38 \\
\bottomrule
\end{tabular}
\caption{Average domain ranks based on Overall scores, with lower ranks indicating easier domains and higher ranks indicating greater difficulty.}
\label{tab:domain_rank}
\end{wraptable}

\paragraph{Performance across Domains}
As shown in Table~\ref{tab:domain}, \agent{} variants remain strong across domains and achieve leading performance in most cases, suggesting good generalization across a diverse set of long-form multimodal research tasks. To characterize domain-level difficulty, we use a ranking-based analysis rather than directly averaging scores, since ranking is less sensitive to unusually high or low values from individual systems. For each system, we rank domains by Overall score and then compute each domain’s average rank across systems. Table~\ref{tab:domain_rank} shows that \textit{History \& Society} and \textit{Education \& Culture} are relatively easier domains, whereas \textit{Technology \& Intelligence}, \textit{Finance \& Business}, and \textit{Law \& Policy} appear more challenging. This pattern is intuitive, as technology, finance, and policy often involve denser terminology, faster-changing factual content, and stricter citation-grounding requirements, all of which increase the difficulty of long-form synthesis and text--visual interleaved report generation. By contrast, domains such as history and culture often allow broader narrative organization and place less pressure on highly technical grounding.

\begin{figure*}[t]
    \centering
    \includegraphics[width=\linewidth]{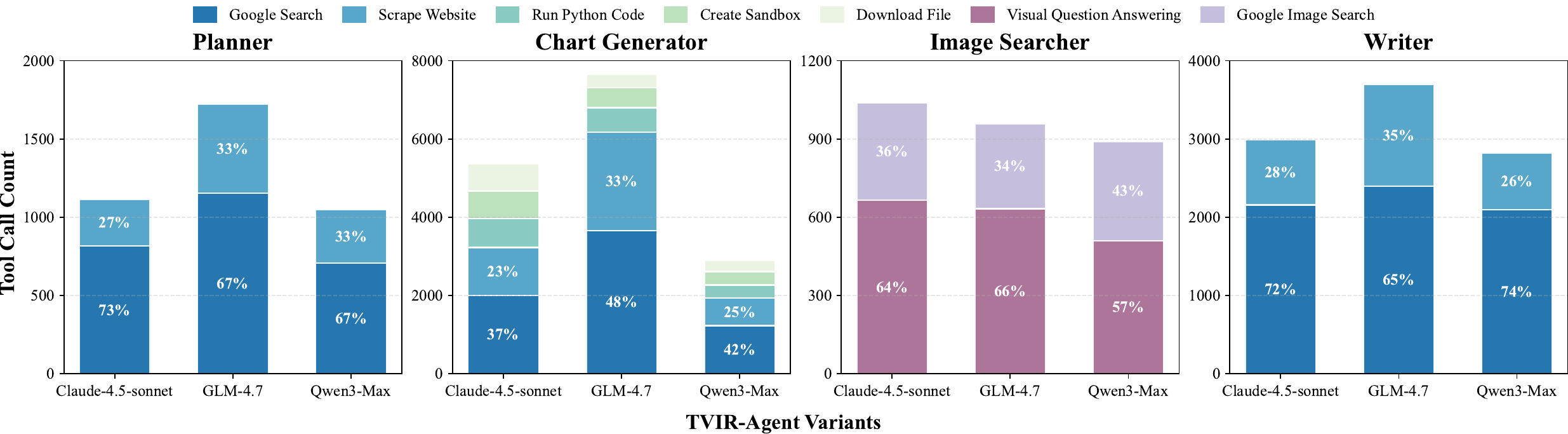}
    \caption{Tool usage distribution of \agent{} variants across major components.}
    \label{fig:tool_usage}
\end{figure*}

\paragraph{Tool Usage Analysis within \agent{}}
Figure~\ref{fig:tool_usage} shows differences in tool usage patterns across \agent{} variants. To quantify the amount of evidence-backed information, we additionally report \textbf{Average Effective Citations (AEC)} in Figure~\ref{fig:citation}, with its definition provided in Appendix~\ref{app:aec}. TVIR-Agent (GLM-4.7) shows the heaviest use of search and scraping tools, especially in the Planner and Chart Generator, which is consistent with its highest AEC of \textbf{102.41}. However, under a limited agent-turn budget, overly extensive retrieval may come at the expense of chart generation: despite planning an average of 8.66 charts per task, it generates only 3.33, yielding a chart fulfillment rate of 38.45\%. By contrast, TVIR-Agent (Claude-4.5-Sonnet) adopts a more balanced tool usage profile, maintaining a relatively high AEC of 86.14 while achieving the highest chart fulfillment rate of \textbf{94.61\%}. Overall, these results suggest that the relative strengths of \agent{} variants depend not only on backbone model capability, but also on how tool usage is allocated between retrieval and chart generation.


\begin{wrapfigure}{r}{0.55\textwidth}
  \includegraphics[width=\linewidth]{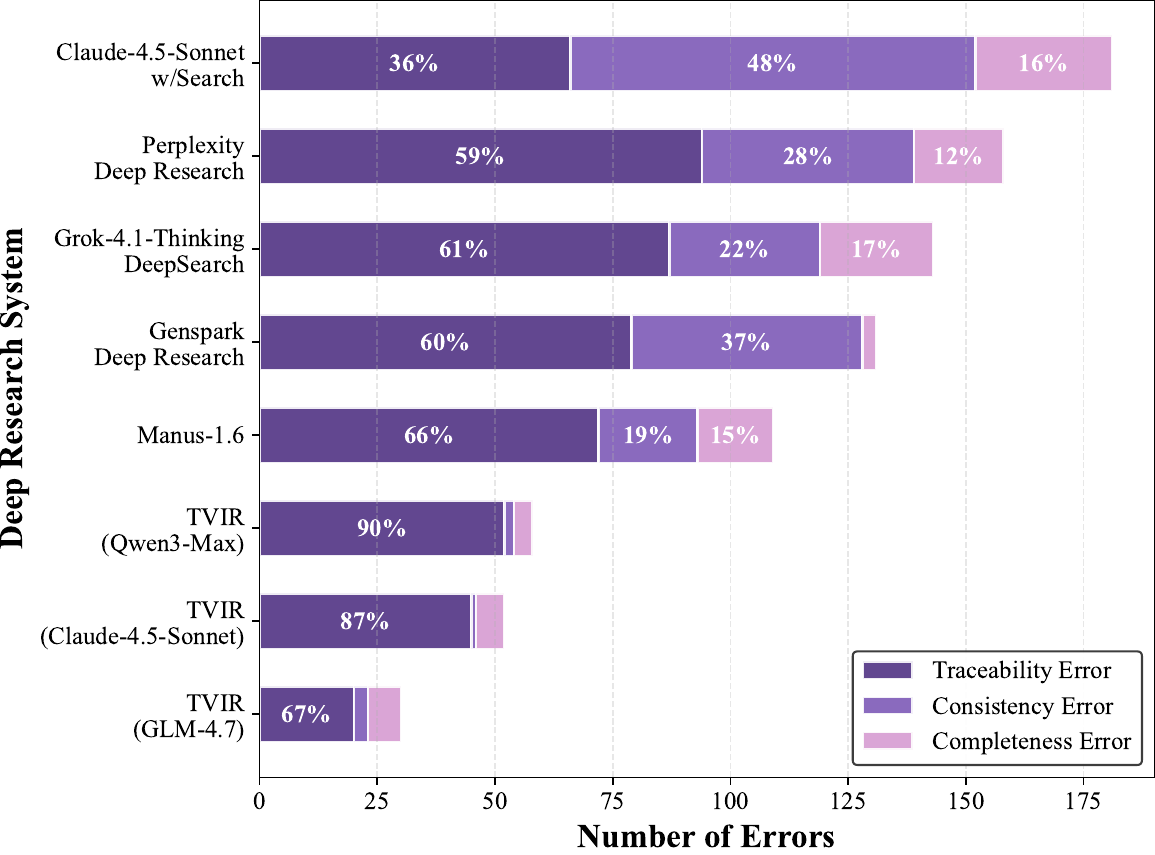}
    \caption{Structural error distributions across evaluated deep research systems.}
    \label{fig:error_case}
\end{wrapfigure}

\paragraph{Structural Error Analysis}
To further examine the reliability of reports, we analyze three types of structural errors: \textbf{Traceability Error}, \textbf{Consistency Error}, and \textbf{Completeness Error}. Traceability Error refers to facts, data, or figures presented without explicit and accessible sources; Consistency Error reflects failures in the internal indexing system, including missing numbers or duplicated entries; Completeness Error captures unusable or incomplete components referenced in the text, such as broken figures or absent captions. As shown in Figure~\ref{fig:error_case}, \agent{} variants produce substantially fewer structural errors than commercial systems overall, suggesting better end-to-end control over citation management and multimodal report assembly, with TVIR-Agent (GLM-4.7) achieving the lowest total error count. Nevertheless, traceability remains a persistent challenge across systems, and even \agent{} exhibits a nontrivial number of errors.

\subsection{Validation of the Evaluation Framework}

\paragraph{Reliability of Information Extraction}
To validate the reliability of the preprocessing stage, we manually annotate ground truth for 90 reports in total, including 10 reports per system across 9 systems. The annotation covers three extraction targets: (i) reference entries, (ii) fact--citation pairs, and (iii) figure elements. As shown in Table~\ref{tab:extraction_quality}, LLM-based extraction achieves near-perfect precision, recall, and F1 across all three targets, suggesting that this stage introduces negligible noise into downstream evaluation. In addition, since metrics such as Citation Support and Chart--Source Consistency depend on webpage content rather than URLs alone, we retrieve webpage text via the Serper API. Across the full evaluation set of 9 $\times$ 100 reports, the retrieval success rate is \textbf{96.53\%}, including cases where the source webpage itself is no longer available, indicating limited impact on evaluation coverage.

\begin{table}[h]
\centering
\begin{tabular}{lccc}
\toprule
\textbf{Extraction Target} & \textbf{Precision} & \textbf{Recall} & \textbf{F1} \\
\midrule
Reference entries   & 100.00 & 100.00 & 100.00 \\
Fact--citation pairs & 99.55  & 99.20  & 99.35  \\
Figure elements     & 100.00 & 100.00 & 100.00 \\
\bottomrule
\end{tabular}
\caption{Performance of LLM-based information extraction.}
\label{tab:extraction_quality}
\end{table}

\paragraph{Human Alignment}
To assess the reliability of our automated evaluation, we conduct a large-scale human alignment study on reports from 8 systems over the full benchmark of 100 tasks, excluding Gemini-3-Pro Deep Research because it generates text-only reports, making VA inapplicable and Overall unavailable. We recruit 20 annotators with Master's degrees and relevant domain expertise, and each report is independently evaluated by three annotators. Following DeepResearch Bench~\citep{du2025deepresearchbenchcomprehensivebenchmark}, we adopt \textbf{Pairwise Agreement Rate (PAR)} and \textbf{Overall Pearson Correlation (OPC)} as the primary metrics, and further report \textbf{Overall Spearman Correlation (OSC)} to capture rank-based consistency. See Appendix~\ref{app:agreement_metrics} for metric definitions. Table~\ref{tab:judge_consistency} shows strong agreement between automated evaluation and human judgments. We also report human inter-annotator agreement, with an Overall PAR of \textbf{74.20}.

\paragraph{Robustness Across Judge LLMs}
To further assess the robustness of the evaluation framework, we compare the results produced by two judge LLMs, GPT-5.2 and Gemini-2.5-Pro~\citep{comanici2025gemini25pushingfrontier}. Table~\ref{tab:cross_judge_consistency} shows high agreement across TA, VA, and Overall. In particular, OPC remains near-perfect in all three dimensions, while OSC and PAR are also consistently strong. These results provide strong evidence that the evaluation framework is robust to the choice of judge LLM.

\begin{table}[h]
    \centering
    \begin{minipage}{0.45\textwidth}
        \centering
        \begin{tabular}{lccc}
        \toprule
        \textbf{Dimension} & \textbf{PAR} & \textbf{OPC} & \textbf{OSC} \\
        \midrule
        TA & 70.00 & 99.12 & 97.62 \\
        VA & 81.07 & 99.42 & 100.00 \\
        \midrule
        Overall & 78.39 & 99.73 & 97.62 \\
        \bottomrule
        \end{tabular}
        \caption{Agreement between automated evaluation and human judgments.}
        \label{tab:judge_consistency}
    \end{minipage}
    \hfill
    \begin{minipage}{0.45\textwidth}
        \centering
        \begin{tabular}{lccc}
        \toprule
        \textbf{Dimension} & \textbf{PAR} & \textbf{OPC} & \textbf{OSC} \\
        \midrule
        TA      & 81.96 & 99.42 & 92.86  \\
        VA      & 88.04 & 99.26 & 100.00 \\
        \midrule
        Overall & 86.07 & 99.44 & 97.62  \\
        \bottomrule
        \end{tabular}
        \caption{Agreement between the two judge LLMs, GPT-5.2 and Gemini-2.5-Pro.}
        \label{tab:cross_judge_consistency}
    \end{minipage}
\end{table}

\subsection{Ablation Studies}
\label{sec:ablation}

To examine the contribution of key components in \agent{}, we conduct ablation studies using TVIR-Agent (Claude-4.5-Sonnet) as the reference system. Specifically, we remove \textbf{research notes}, the \textbf{Image Searcher}, and the \textbf{Chart Generator}, respectively, while keeping the rest of the pipeline unchanged. As shown in Table~\ref{tab:ablation_main}, removing any component leads to a drop in overall performance. Among them, removing the Chart Generator has the largest effect, reducing VA from 78.62 to 60.91 and Overall from 73.92 to 63.84, which highlights its central role in visual synthesis and cross-modal alignment. Removing the Image Searcher leads to a clear decline across all metrics, whereas the effect of removing the research notes is relatively small. These results show that the performance gains of \agent{} come from the complementary contributions of research-grounded planning and specialized visual capabilities.

\begin{table}[htbp]
\centering
\begin{tabular}{lccc}
\toprule
\textbf{System Variant} & \textbf{TA} & \textbf{VA} & \textbf{Overall} \\
\midrule
TVIR-Agent & 69.23 & 78.62 & 73.92 \\
\midrule
w/o research notes & 68.63 & 78.42 & 73.52 \\
w/o Image Searcher & 67.82 & 77.23 & 72.53 \\
w/o Chart Generator & 66.77 & 60.91 & 63.84 \\
\bottomrule
\end{tabular}
\caption{Ablation results for TVIR-Agent (Claude-4.5-Sonnet) on 20 randomly sampled tasks.}
\label{tab:ablation_main}
\end{table}

\section{Conclusion}

We introduce TVIR, a unified benchmark and agentic framework for text--visual interleaved report generation. It includes \bench{}, an expert-curated benchmark with a dual-path evaluation framework covering both textual and visual assessment, and \agent{}, a hierarchical multi-agent framework that explicitly models visual evidence throughout planning and writing. Extensive experiments show that \agent{} achieves strong overall performance and improves evidence grounding and cross-modal alignment over existing systems, while also revealing a key limitation of current deep research systems: they remain much stronger at textual synthesis than at integrating visual assets. We hope TVIR will provide a foundation for future work on trustworthy multimodal deep research agents. 
\bibliographystyle{unsrtnat}
\bibliography{main} 

\appendix
\section{Data Design Details}
\label{app:a}
\subsection{Definitions of Task Design Principles}
\label{app:task_design}

\paragraph{Role-Driven}
Each task should be centered on a clearly identifiable user with a specific professional identity or functional role, such as a policymaker, clinician, investment analyst, researcher, or engineer. The role should be concrete and representative of a real user group that may plausibly raise such a research need in practice, rather than a vague or generic subject such as "a normal user". To strengthen realism and task specificity, the task context should also describe the user’s objective, constraints, or decision scenario. For example, instead of asking a general question about a biomedical trend, a task may be framed from the perspective of a head of R\&D at a biopharmaceutical company who needs to assess the clinical and commercial prospects of a new therapeutic direction. This design principle ensures that tasks are anchored in authentic use cases rather than abstract knowledge queries.

\paragraph{Demand-Oriented } 
Each task should express a clear and structured research demand within a specific domain or topic. A task may include multiple sub-questions, but they should remain logically connected and confined to the same thematic scope rather than being artificially assembled from unrelated domains. Each sub-question should correspond to a well-defined information goal, avoiding open-ended prompts such as "give your opinion" without substantive constraints. We encourage tasks to be presented in a structured manner, for example by explicitly listing the aspects that need to be covered, so that the model has a clear pathway for organizing its response and the evaluator has a clear basis for judging completeness. While complex tasks are allowed, the overall instruction should remain coherent, bounded, and operational, without ambiguity, redundancy, or internal contradiction.

\paragraph{Deep Research}
Tasks should require substantially more than surface-level information aggregation. They should encourage the model to synthesize evidence from multiple sources, construct causal or explanatory chains, compare competing claims, and produce conclusions or recommendations. Relevant evidence may include academic findings, statistical data, case materials, or expert viewpoints, and the task should implicitly or explicitly require the model to explain how such evidence supports its claims. In addition, tasks should encourage critical analysis by asking the model to recognize conflicts across sources, limitations in data, uncertainty in interpretation, or weaknesses in methods and assumptions. A well-designed deep research task should therefore form a closed analytical loop: evidence gathering, reasoning and critique, and then conclusion or actionable recommendation.

\paragraph{Frontier-Focused}
Tasks should target recent developments, emerging challenges, and open questions across domains, so as to evaluate whether models can engage with up-to-date knowledge rather than relying mainly on static background information. In general, topics are expected to focus on new technologies, policy changes, market shifts, scientific breakthroughs, or other developments that have become salient within the past two to three years. Tasks are encouraged to rely on recent and authoritative sources, especially materials published in or after 2024, such as top-tier conference papers, industry white papers, government releases, company filings, or official statistical reports. When a task involves trend analysis or quantitative comparison, it should call for the latest available data, ideally from 2024--2026 when accessible, with clear attention to timestamps. For frontier topics where no stable consensus yet exists, the task may ask the model to present and compare different institutional or scholarly viewpoints and analyze the basis of their disagreement.

\paragraph{Multimodal Integration }
Tasks should explicitly require the integration and presentation of multimodal information, so that visual elements become a meaningful part of the research process rather than decorative additions. Such elements may include publicly retrievable images from the web, such as model architecture diagrams, algorithm workflow illustrations, historical portraits, environmental photographs, or scene images, as well as code-generated charts based on real, publicly accessible data, such as line charts, radar charts, heatmaps, or pie charts. In task wording, multimodal needs should usually be expressed naturally through verbs and phrases such as "visualize", "plot", "show", "illustrate", or "provide a figure with explanation", without always explicitly specifying whether the model must retrieve an existing image or generate a chart. Instead, the intended visual type should be implied by the request itself, for example by asking for a radar chart to compare alternatives or an architecture diagram to explain a technical system. All multimodal elements must directly serve the core analytical objective and should be added only when they improve interpretation, explanation, or decision support. For example, a heatmap may be used to show regional differences in policy outcomes, an architecture diagram to clarify the innovation of a model design, or a time-series plot to support claims about market growth. To ensure practical feasibility, any required retrieved image must be publicly available online, and any required generated chart must rely on real and publicly accessible data sources.

\subsection{Definitions of Task Complexity Levels}
\label{app:complexity_levels}

\paragraph{Low Complexity}
Low-complexity tasks are relatively concise, typically around 200 Chinese characters or about 130 English words. They usually contain \textbf{one to three multimodal requirements} and are primarily intended to assess whether a system can handle focused deep research requests with limited structural and coordination burden. Such tasks may be written in a compact paragraph or brief outline form, and explicit bullet-point decomposition is not required.

\paragraph{Medium Complexity}
Medium-complexity tasks are more elaborate, typically around 400 Chinese characters or about 260 English words, and contain \textbf{two to four multimodal requirements}. They are expected to involve a clearer internal structure and usually benefit from being organized into \textbf{three to four major points}, each with \textbf{one to three sub-questions}. These tasks impose greater demands on instruction following, cross-source synthesis, and the coordinated use of text and visual evidence.

\paragraph{High Complexity}
High-complexity tasks are the most demanding, typically around 600 Chinese characters or about 390 English words, and contain \textbf{three to five multimodal requirements}. They must be presented in an explicitly structured format, usually with \textbf{four to five major points}, each containing \textbf{two to four sub-questions}. These tasks place substantial demands on long-horizon planning, detailed instruction tracking, and multimodal integration across multiple analytical dimensions.

\subsection{Definitions of High-Level Functional Types}
\label{app:functional_types}

To guide task design, we predefine eight high-level functional types that specify the intended analytical role of sub-questions in \bench{}. Each sub-question is designed to primarily instantiate one of these functions, and Table~\ref{tab:functional_types} gives their definitions.

\begin{table*}[h]
\centering
\setlength{\tabcolsep}{6pt}
\renewcommand{\arraystretch}{1.2}
\begin{tabular}{p{0.26\textwidth} p{0.68\textwidth}}
\toprule
\textbf{Functional Type} & \textbf{Definition} \\
\midrule

\textbf{Status Characterization} 
& Questions that describe the current state, structure, characteristics, or distribution of an object, issue, or system. \\

\textbf{Comparative Analysis} 
& Questions that compare two or more entities to identify similarities, differences, relative strengths and weaknesses, or rankings. \\

\textbf{Mechanism Explanation} 
& Questions that explain the causes, mechanisms, or underlying principles behind a phenomenon, outcome, or behavior. \\

\textbf{Evolution Tracing} 
& Questions that examine the origin and development of an object, idea, technology, or institution over time. \\

\textbf{Impact Assessment} 
& Questions that evaluate the effects, consequences, benefits, risks, or limitations of a policy, technology, event, or intervention. \\

\textbf{Trend Prediction} 
& Questions that infer likely future developments or outcomes based on current data, patterns, or models. \\

\textbf{Foundational Exploration} 
& Questions that probe fundamental or frontier issues to investigate principles, unknown mechanisms, or the nature of a problem. \\

\textbf{Solution Construction} 
& Questions that require designing a framework, model, system, strategy, or actionable plan to solve a problem or achieve a goal. \\

\bottomrule
\end{tabular}
\caption{Definitions of the eight high-level functional types used in \bench{}.}
\label{tab:functional_types}
\end{table*}
\section{Fine-Grained Evaluation Metrics Computation Details}
\label{app:fine_grained_metrics}

All fine-grained metrics are first computed on a normalized scale in $[0,1]$ and then linearly rescaled to $[0,100]$ for presentation.

\subsection{Textual Assessment Metrics}
\label{app:textual_metrics}

Let a report contain $M$ extracted fact--citation pairs and $K$ checklist items. For rubric-based report-level metrics, let $D$ denote the number of scoring dimensions in the corresponding rubric.

\paragraph{Citation Support}
For each extracted fact--citation pair, the judge assigns
\[
s_i^{\mathrm{CS}} \in \{1,\;0.5,\;0\},
\]
corresponding to \textit{supported}, \textit{partially supported}, and \textit{unsupported}, respectively. The Citation Support score is
\[
\mathrm{CS}
=
\frac{1}{M}
\sum_{i=1}^{M}
s_i^{\mathrm{CS}}.
\]
If no fact--citation pair is available, the score is set to $0$.

\paragraph{Instruction Alignment}
For each checklist item, the judge assigns
\[
s_j^{\mathrm{IA}} \in \{1,\;0.5,\;0\},
\]
corresponding to \textit{fully satisfied}, \textit{partially satisfied}, and \textit{not satisfied}, respectively. The Instruction Alignment score is
\[
\mathrm{IA}
=
\frac{1}{K}
\sum_{j=1}^{K}
s_j^{\mathrm{IA}}.
\]

\paragraph{Writing Quality}
Writing Quality is assessed using a report-level rubric with $D_{\mathrm{WQ}}=4$ dimensions. Let
\[
r_d^{\mathrm{WQ}} \in \{1,2,\dots,10\}
\]
be the judge score for dimension $d$. The normalized score is
\[
\mathrm{WQ}
=
\frac{1}{10D_{\mathrm{WQ}}}
\sum_{d=1}^{D_{\mathrm{WQ}}}
r_d^{\mathrm{WQ}}.
\]

\paragraph{Analytical Depth \& Breadth}
Analytical Depth \& Breadth is assessed using a report-level rubric with $D_{\mathrm{ADB}}=5$ dimensions. Let
\[
r_d^{\mathrm{ADB}} \in \{1,2,\dots,10\}
\]
be the judge score for dimension $d$. The normalized score is
\[
\mathrm{ADB}
=
\frac{1}{10D_{\mathrm{ADB}}}
\sum_{d=1}^{D_{\mathrm{ADB}}}
r_d^{\mathrm{ADB}}.
\]

\paragraph{Factual \& Logical Consistency}
Let $N_{\mathrm{issues}}$ denote the number of distinct internal contradiction issues identified by the judge. This issue count is mapped to a discrete 10-point score $r^{\mathrm{FLC}}$, which is then normalized as
\[
\mathrm{FLC}
=
\frac{r^{\mathrm{FLC}}}{10}.
\]

\begin{table}[h]
\centering
\begin{tabular}{cc}
\toprule
$N_{\mathrm{issues}}$ & $r^{\mathrm{FLC}}$ \\
\midrule
$0$ & $10$ \\
$1$--$2$ & $9$ \\
$3$--$4$ & $8$ \\
$5$--$6$ & $7$ \\
$7$--$8$ & $6$ \\
$9$--$10$ & $5$ \\
$11$--$12$ & $4$ \\
$13$--$14$ & $3$ \\
$15$--$17$ & $2$ \\
$\geq 18$ & $1$ \\
\bottomrule
\end{tabular}
\caption{Mapping from contradiction count to the FLC score.}
\label{tab:flc_mapping}
\end{table}

\subsection{Visual Assessment Metrics}
\label{app:visual_metrics}

Let a report contain $N_{\mathrm{img}}$ extracted images, $N_{\mathrm{chart}}$ extracted charts, and $N_{\mathrm{fig}}$ extracted figure elements in total.

\paragraph{Multimodal Composition}
Multimodal Composition is assessed using a report-level rubric with $D_{\mathrm{MC}}=2$ dimensions. Let
\[
r_d^{\mathrm{MC}} \in \{1,2,\dots,10\}
\]
be the judge score for dimension $d$. The normalized score is
\[
\mathrm{MC}
=
\frac{1}{10D_{\mathrm{MC}}}
\sum_{d=1}^{D_{\mathrm{MC}}}
r_d^{\mathrm{MC}}.
\]
If no figure element is present, the score is set to $0$.

\paragraph{Figure Quality}
Figure Quality combines \textbf{Image Quality (IQ)} and \textbf{Chart Quality (CQ)}.

For each image $i$, we compute four normalized sub-scores: resolution $R_i$, aspect ratio $A_i$, sharpness $S_i$, and contrast $C_i$. The raw image quality score is defined as
\[
Q_i = 0.25R_i + 0.15A_i + 0.35S_i + 0.25C_i.
\]
The weights are human-calibrated rather than manually assigned. Expert annotators rated extracted images on the same four sub-dimensions and an overall image quality score, each ranging from 1 to 10. We then fit a linear regression model from the four sub-scores to the overall rating and normalized the fitted coefficients to obtain the final weights. To penalize near-duplicate images, we compute a duplicate penalty $p_{\mathrm{dup}}$ based on perceptual-hash similarity. The Image Quality score is
\[
\mathrm{IQ}
=
\left(
\frac{1}{N_{\mathrm{img}}}
\sum_{i=1}^{N_{\mathrm{img}}}
Q_i
\right)
(1-p_{\mathrm{dup}}).
\]
If no image is present, the score is set to $0$.

For each chart $j$, the judge evaluates a binary checklist with $D_{\mathrm{CQ}}=10$ items. Let
\[
c_{j,d} \in \{0,1\}
\]
be the score for checklist item $d$. The Chart Quality score is
\[
\mathrm{CQ}
=
\frac{1}{D_{\mathrm{CQ}}N_{\mathrm{chart}}}
\sum_{j=1}^{N_{\mathrm{chart}}}
\sum_{d=1}^{D_{\mathrm{CQ}}}
c_{j,d}.
\]
If no chart is present, the score is set to $0$.

The final Figure Quality score is the count-weighted average of IQ and CQ:
\[
\mathrm{FQ}
=
\frac{
N_{\mathrm{img}}\cdot \mathrm{IQ}
+
N_{\mathrm{chart}}\cdot \mathrm{CQ}
}{
N_{\mathrm{img}} + N_{\mathrm{chart}}
}.
\]
If no figure element is present, the score is set to $0$.

\paragraph{Figure Caption Quality}
For each figure element $k$, the judge assigns 1--10 scores on $D_{\mathrm{FCQ}}=3$ dimensions. Let
\[
r_{k,d}^{\mathrm{FCQ}} \in \{1,2,\dots,10\}
\]
be the dimension score. The normalized per-figure score is
\[
q_k^{\mathrm{FCQ}}
=
\frac{1}{10D_{\mathrm{FCQ}}}
\sum_{d=1}^{D_{\mathrm{FCQ}}}
r_{k,d}^{\mathrm{FCQ}}.
\]
The report-level Figure Caption Quality score is
\[
\mathrm{FCQ}
=
\frac{1}{N_{\mathrm{fig}}}
\sum_{k=1}^{N_{\mathrm{fig}}}
q_k^{\mathrm{FCQ}}.
\]
If no figure element is present, the score is set to $0$. In our implementation, missing captions are assigned a score of $0$.

\paragraph{Figure--Context Integration}
For each figure element $k$, the judge assigns 1--10 scores on $D_{\mathrm{FCI}}=3$ dimensions. Let
\[
r_{k,d}^{\mathrm{FCI}} \in \{1,2,\dots,10\}
\]
be the score for dimension $d$. The normalized per-figure score is
\[
q_k^{\mathrm{FCI}}
=
\frac{1}{10D_{\mathrm{FCI}}}
\sum_{d=1}^{D_{\mathrm{FCI}}}
r_{k,d}^{\mathrm{FCI}}.
\]
The report-level Figure--Context Integration score is
\[
\mathrm{FCI}
=
\frac{1}{N_{\mathrm{fig}}}
\sum_{k=1}^{N_{\mathrm{fig}}}
q_k^{\mathrm{FCI}}.
\]
If no figure element is present, the score is set to $0$. In our implementation, figures with missing context are assigned a score of $0$.

\paragraph{Chart--Source Consistency}
For each chart $j$, let $N_j^{\mathrm{CSC}}$ denote the number of distinct contradiction issues identified by the judge. This issue count is mapped to a discrete 10-point score $r_j^{\mathrm{CSC}}$, which is then normalized as
\[
\mathrm{CSC}
=
\frac{1}{10N_{\mathrm{chart}}}
\sum_{j=1}^{N_{\mathrm{chart}}}
r_j^{\mathrm{CSC}}.
\]
If no chart is present, the score is set to $0$. In our implementation, charts without usable cited references are assigned a score of $0$.

\begin{table}[h]
\centering
\begin{tabular}{cc}
\toprule
$N_j^{\mathrm{CSC}}$ & $r_j^{\mathrm{CSC}}$ \\
\midrule
$0$ & $10$ \\
$1$ & $9$ \\
$2$ & $8$ \\
$3$ & $7$ \\
$4$ & $6$ \\
$5$ & $5$ \\
$6$ & $4$ \\
$7$ & $3$ \\
$8$ & $2$ \\
$\geq 9$ & $1$ \\
\bottomrule
\end{tabular}
\caption{Mapping from contradiction count to the CSC score.}
\label{tab:csc_mapping}
\end{table}

\section{Evaluation Implementation Details}
\label{app:evaluation_implementation}

Since current LLMs are still not reliable at processing long multimodal reports in a single pass, especially when lengthy text and many figures must be considered jointly, such evaluation can easily introduce hallucinations. Therefore, the two report-level metrics that involve multimodal elements, \textbf{Instruction Alignment} and \textbf{Multimodal Composition}, are evaluated based on the textual report representation. In this setting, the judge mainly relies on figure captions and insertion markers to determine whether multimodal requirements are satisfied and how multimodal elements are organized across the report.

This design creates two practical issues. First, invalid visual references, such as missing figures, broken links, corrupted files, empty paths, or pseudo-visual content (e.g., ASCII diagrams, Mermaid-style text graphics, and other non-rendered textual substitutes), may misleadingly appear as multimodal evidence in the report text. To avoid this, we preprocess reports by replacing the URL or local path of any unusable visual element with an empty value, so that it is treated as invalid in later evaluation stages.

Second, figure captions may be fluent but semantically inconsistent with the actual figure, which can also distort text-based judgments. To mitigate this problem, we perform a caption-revision step using the \textbf{Figure Caption Quality} evaluation, where the judge has direct access to the figure itself. If any scoring dimension of a caption is at most $6$, the caption is treated as unreliable and replaced with the judge-generated revision from the same step; if a figure has no caption, the generated revision is inserted when possible. The resulting corrected report is then used for subsequent evaluation, providing a cleaner and more faithful textual representation of the report's actual multimodal content.

\section{Experimental Implementation Details}
\label{app:experimental_implementation}

\subsection{Report Collection}
\label{app:report_collection}

For the six commercial closed-source systems, reports are collected through their official web interfaces rather than APIs. To guide these systems to produce reports in the text--visual interleaved format required by our benchmark, we use a unified system prompt for all of them, as provided in Appendix~\ref{app:report_generation_prompts}. This prompt is largely shared across systems, with only two practical adjustments: the requested output format is specified as either \textbf{Markdown} or \textbf{HTML}, depending on the platform's native support, and the chart-generation instruction is adapted to system capability, as some systems are better suited to Python-based chart generation while others align more naturally with JavaScript-based rendering. Although \textbf{Gemini-3-Pro Deep Research} can only produce text-only reports in our setup, we still provide it with the same multimodal report-generation prompt for fairness and consistency.

Among the evaluated commercial systems, \textbf{Claude-4.5-Sonnet w/Search} and \textbf{Genspark Deep Research} produce deep research reports only in HTML format. Since our evaluation framework takes Markdown reports as input, these outputs are converted into Markdown before preprocessing and scoring. During conversion, we strictly verify that: (1) all textual content is preserved completely; (2) citations, source attributions, and provenance information remain accurate and aligned with the original output; and (3) all non-textual contents, including charts and retrieved images, are saved as local files and inserted at the correct positions in the Markdown report. For systems that can directly generate Markdown reports, no format conversion is required. However, all reports, including those generated by our \agent{} variants, are normalized into the same storage format for downstream evaluation.

\begin{table*}[t]
  \centering
  \resizebox{\textwidth}{!}{
  \begin{tabular}{l ccccc ccccc ccc}
    \toprule
    \multirow{2}{*}{\textbf{System}} 
    & \multicolumn{5}{c}{\textbf{Textual Assessment}} 
    & \multicolumn{5}{c}{\textbf{Visual Assessment}} 
    & \multicolumn{3}{c}{\textbf{Aggregate}} \\
    \cmidrule(lr){2-6} \cmidrule(lr){7-11} \cmidrule(lr){12-14}
    \cmidrule(lr){2-6} \cmidrule(lr){7-11} \cmidrule(lr){12-14}
    & CS & IA & WQ & ADB & FLC
    & FQ & MC & FCQ & FCI & CSC
    & \textbf{TA} & \textbf{VA} & \textbf{Overall} \\
    \midrule
    \multicolumn{14}{c}{\textbf{Commercial Closed-Source Systems}}\\
    \midrule
    Gemini-3-Pro Deep Research & 0.002 & 0.002 & 0.003 & 0.004 & 0.003 & - & - & - & - & - & 0.002 & - & -\\
    Grok-4.1-Thinking DeepSearch & 0.006 & 0.002 & 0.002 & 0.004 & 0.000 & 0.001 & 0.007 & 0.005 & 0.004 & 0.000 & 0.002 & 0.002 & 0.000\\
    Claude-4.5-Sonnet w/Search & 0.002 & 0.011 & 0.002 & 0.008 & 0.014 & 0.003 & 0.020 & 0.003 & 0.004 & 0.007 & 0.002 & 0.006 & 0.002\\
    Perplexity Deep Research & 0.002 & 0.003 & 0.007 & 0.006 & 0.006 & 0.000 & 0.005 & 0.001 & 0.004 & 0.003 & 0.005 & 0.001 & 0.003\\
    Genspark Deep Research & 0.003 & 0.010 & 0.001 & 0.006 & 0.003 & 0.002 & 0.008 & 0.004 & 0.002 & 0.014 & 0.001 & 0.004 & 0.002\\
    Manus-1.6 & 0.006 & 0.005 & 0.002 & 0.004 & 0.005 & 0.005 & 0.007 & 0.002 & 0.001 & 0.001 & 0.003 & 0.000 & 0.001\\
    \midrule
    \multicolumn{14}{c}{\textbf{\agent{} Variants (Ours)}}\\
    \midrule
    TVIR-Agent (Qwen3-Max) & 0.002 & 0.015 & 0.001 & 0.003 & 0.011 & 0.002 & 0.005 & 0.008 & 0.005 & 0.018 & 0.004 & 0.003 & 0.001\\
    TVIR-Agent (GLM-4.7) & 0.003 & 0.006 & 0.002 & 0.006 & 0.009 & 0.001 & 0.011 & 0.004 & 0.006 & 0.010 & 0.002 & 0.004 & 0.003\\
    TVIR-Agent (Claude-4.5-Sonnet) & 0.002 & 0.001 & 0.005 & 0.002 & 0.000 & 0.006 & 0.004 & 0.003 & 0.001 & 0.004 & 0.001 & 0.001 & 0.001\\
    
    \bottomrule
    \end{tabular}
    }
  \caption{Standard deviations of the main results on TVIR-BENCH over three independent runs.}
  \label{tab:std}
\end{table*}

\subsection{Variance Across Independent Runs}

To assess the stability of our evaluation framework, all results are independently evaluated three times under the same setup, and the standard deviation of each metric is reported in Table~\ref{tab:std}. The standard deviations are consistently small across all evaluated systems, indicating strong run-to-run stability. In particular, the aggregate metrics TA, VA, and Overall show very limited variation. These results suggest that the performance differences reported in the main paper are robust and unlikely to be driven by random variation across runs.

\begin{figure*}[t]
    \centering
    \includegraphics[width=\linewidth]{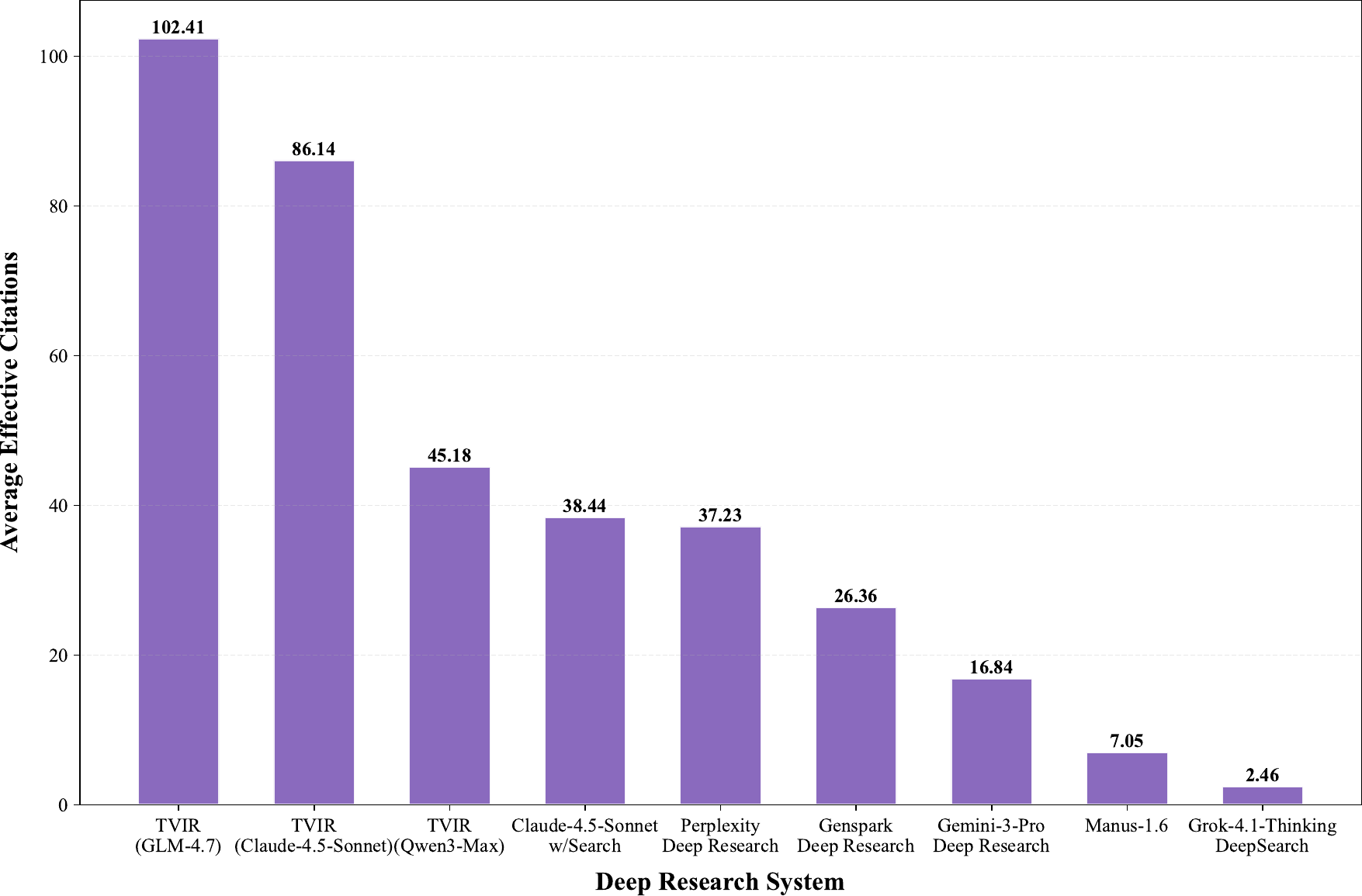}
    \caption{Average Effective Citations (AEC) across evaluated deep research systems.}
    \label{fig:citation}
\end{figure*}

\subsection{Average Effective Citations}
\label{app:aec}

To complement citation-quality evaluation, we additionally report \textbf{Average Effective Citations (AEC)}, which measures how much verifiably supported information a system provides on average for each task. Intuitively, AEC reflects the average quantity of factual statements in generated reports that are supported by their cited references, and therefore captures a system’s effective use of evidence beyond a normalized accuracy score.

Let $T$ denote the set of all tasks, and let $|T|$ be the total number of tasks. For each task $t \in T$, let $U_t$ denote the set of extracted fact--citation pairs after preprocessing. For each pair $i \in U_t$, the Citation Support evaluator assigns a support score
\[
s_{i,t}^{\mathrm{CS}} \in \{1,\;0.5,\;0\},
\]
corresponding to \textit{supported}, \textit{partially supported}, and \textit{unsupported}, respectively. We define the effective citation count for task $t$ as the sum of these support scores over all extracted pairs:
\[
E_t = \sum_{i \in U_t} s_{i,t}^{\mathrm{CS}}.
\]

The \textbf{Average Effective Citations} over the full benchmark is then computed as
\[
\mathrm{AEC}
=
\frac{1}{|T|}
\sum_{t \in T}
E_t
=
\frac{1}{|T|}
\sum_{t \in T}
\sum_{i \in U_t}
s_{i,t}^{\mathrm{CS}}.
\]

Compared with the normalized Citation Support score, which measures the average support quality of extracted factual statements, AEC emphasizes the \emph{absolute amount} of useful, evidence-backed information that a system delivers per task. A system may achieve a high normalized citation-support score by making relatively few but well-supported statements, while AEC further rewards systems that can sustain strong evidence grounding at larger informational scale.

\subsection{Agreement Metrics}
\label{app:agreement_metrics}

\paragraph{Pairwise Agreement Rate}
Pairwise Agreement Rate (PAR) measures how often two evaluators induce the same pairwise preference over reports for the same task. Let $T$ denote the set of tasks, and let $S$ denote the set of evaluated systems. For each task $t \in T$, there are
\[
N_p = \binom{|S|}{2}
\]
unordered report pairs. For a given task $t$ and report pair $p$, let
\[
I(t,p)=
\begin{cases}
1, & \text{if the pairwise preferences agree},\\
0, & \text{otherwise}.
\end{cases}
\]

Here, a pairwise preference is determined by comparing the scores assigned by the two evaluators to the two reports in the pair; equality is treated as a tie. The Pairwise Agreement Rate is then computed as
\[
\mathrm{PAR}
=
\frac{
\sum\limits_{t \in T}
\sum\limits_{p=1}^{N_p}
I(t,p)
}{
|T| \cdot N_p
}.
\]

\paragraph{Overall Pearson Correlation}
Overall Pearson Correlation (OPC) measures the linear correlation between the system-level average scores produced by two evaluators. Let
\[
X = (x_1, x_2, \dots, x_{|S|})
\]
be the vector of average scores assigned by the first evaluator across all sampled tasks for the evaluated systems, and let
\[
Y = (y_1, y_2, \dots, y_{|S|})
\]
be the corresponding vector produced by the second evaluator. The Overall Pearson Correlation is defined as
\[
\mathrm{OPC} = r_{\mathrm{Pearson}}(X, Y),
\]
where $r_{\mathrm{Pearson}}(\cdot,\cdot)$ denotes the standard Pearson correlation coefficient. This metric evaluates whether the two evaluators assign similar absolute score patterns across systems.

\paragraph{Overall Spearman Correlation}
Overall Spearman Correlation (OSC) measures the rank-based correlation between the system-level average scores produced by two evaluators. Using the same vectors $X$ and $Y$, the Overall Spearman Correlation is defined as
\[
\mathrm{OSC} = r_{\mathrm{Spearman}}(X, Y),
\]
where $r_{\mathrm{Spearman}}(\cdot,\cdot)$ denotes the standard Spearman rank correlation coefficient. Compared with OPC, OSC focuses on whether the two evaluators induce similar system rankings rather than similar absolute score values.

\section{Prompt Templates}
\subsection{Prompt for LLM-Based Task Drafting}
\label{app:task_drafting_prompt}
\begin{promptbox}{Prompt for LLM-Based Task Drafting}

\textbf{SYSTEM PROMPT}
\vspace{0.5em}

\textbf{Project Background}
\vspace{0.5em}

Deep Research refers to a class of AI research assistants or toolchains that automate the full pipeline of ``information retrieval $\rightarrow$ analytical synthesis $\rightarrow$ structured writing,'' aiming to produce comprehensive reports approaching professional analyst quality within a relatively short time. Its typical workflow is as follows: after a user provides a research topic together with requirements on depth and scope, the agent autonomously performs multi-step web search, collects and analyzes a large number of information sources (including text, images, PDFs, and other modalities), dynamically adjusts retrieval strategies, and finally generates a structured research report with citations attached to key claims for verification.

Current mainstream Deep Research benchmarks mainly focus on the generation quality of purely textual reports, typically evaluating outputs such as literature reviews, issue analyses, or conclusion-oriented essays. However, in real-world deep research scenarios---including scientific research, policy analysis, and business intelligence---high-quality reports often rely heavily on the coordinated expression of multimodal information. For example, charts may be used to present key data trends, schematic diagrams to explain complex mechanisms, and images to provide evidence for experimental phenomena. These visual elements are not merely decorative additions; rather, they are deeply coupled with the text at the levels of semantics, logic, and argumentative structure, jointly forming a complete research narrative.

Because existing evaluation frameworks largely ignore the multimodal dimension, they suffer from clear limitations. On the one hand, they cannot assess whether a model can effectively introduce external visual evidence or invoke tools to generate appropriate charts in response to a text-only input, in order to support its analysis and conclusions. On the other hand, they also fail to adequately measure a model's overall performance in terms of image--text consistency, cross-modal logical alignment, and informational complementarity. This evaluation gap may lead models, in practical applications, to produce reports that appear plausible yet lack empirical support, or reports in which visual and textual components are poorly aligned, thereby weakening their credibility and practical utility in real research tasks.

Accordingly, there is an urgent need to construct a Deep Research benchmark for multimodal report generation. Under the setting where the original user input remains textual, such a benchmark should systematically cover multiple dimensions, including the quality of textual generation, the appropriateness of citing external images, the correctness of tool-generated charts, and the semantic coordination and logical coherence between text and visuals. In this way, it can more comprehensively and realistically reflect a model's overall capability in complex research tasks. This project aims to fill that gap and provide standardized, scalable infrastructure for the development and evaluation of next-generation multimodal Deep Research models.

\vspace{0.5em}
\textbf{Task Design Principles}
\vspace{0.5em}

\textbf{1. Role-Driven}

Each task should be centered on a \textbf{clearly identified user with a concrete role or professional identity} (e.g., policymaker, clinician, investment analyst, scientist, student), reflecting authentic research needs and decision contexts in a specific domain.

\begin{itemize}
    \item Each task must explicitly specify the user's role (e.g., ``Head of R\&D at a biopharmaceutical company'').
    \item The user role should be representative of a real professional group that could plausibly raise such a research question in practice.
    \item The task context should include the user's objective, constraints, or decision scenario to enhance realism and relevance.
    \item Avoid generic users such as ``ordinary user'' or ``researcher''; the role should be concrete and clearly recognizable.
\end{itemize}

\textbf{2. Demand-Oriented}

Each task should pose a clear and structured research need within a specific domain. It \textbf{may include multiple related sub-questions}, provided that these are operational and help the model organize its output in a way that supports subsequent evaluation.

\begin{itemize}
    \item A task may include multiple sub-questions, but they must remain within the same domain or theme and should not be artificially stitched together across unrelated fields.
    \item Each sub-question should have a clear informational target; avoid open-ended prompts such as ``What do you think?''
    \item Structured task formulations are encouraged, for example by listing aspects that must be covered, so as to provide a clear path for generation.
    \item Complex tasks are allowed, but the overall logic must remain coherent, the boundaries clear, and the instructions free of ambiguity, redundancy, or contradiction.
\end{itemize}

\textbf{3. Deep Research}

Tasks should require the model to go beyond superficial aggregation of facts, instead demanding multi-source synthesis, causal reasoning, evidence-based argumentation, and critical analysis, thereby demonstrating deeper understanding and integration.

\begin{itemize}
    \item Tasks should go beyond fact listing and require the model to establish logical chains of reasoning.
    \item The use of multi-source evidence is encouraged, including literature findings, statistical data, case images, and expert views, together with explanations of how each source supports the argument.
    \item The model should be asked to identify information conflicts, data limitations, or methodological weaknesses, thereby demonstrating critical thinking.
    \item The output should include clear conclusions or recommendations that close the loop with the preceding analysis.
\end{itemize}

\textbf{4. Frontier-Focused}

Task topics should cover recent developments and key challenges across domains, encouraging the model to use up-to-date literature, data, and multimodal resources that reflect current frontier knowledge.

\begin{itemize}
    \item Topics should \textbf{focus on new technologies, new policies, emerging markets, or scientific breakthroughs that have appeared within the past 2--3 years}.
    \item It is encouraged to \textbf{cite authoritative sources from 2023 onward}, such as top-conference papers, industry white papers, government releases, or public company filings.
    \item If a task involves technical or market trends, it should require the use of the latest available data (e.g., 2024--2025 statistics) with explicit timestamps.
    \item For frontier issues where no consensus yet exists, the task may require the presentation of different schools of thought or institutional positions, together with analysis of their supporting evidence.
\end{itemize}

\textbf{5. Multimodal Integration}

Tasks should explicitly require the integration and presentation of multimodal information, asking the model to combine text, images, charts, and other modalities during analysis in order to improve clarity, explanatory power, and decision support.

\begin{itemize}
    \item Multimodal elements may include: images directly \textbf{retrieved from the web} (e.g., model architecture diagrams, algorithm flowcharts, portraits of historical figures, environmental or scene photographs), or visualizations \textbf{generated by code} based on web-retrieved data (e.g., line charts, radar charts, heatmaps, pie charts).
    \item The task description should naturally embed keywords such as ``plot,'' ``show,'' ``visualize,'' and ``include figure explanations'' so as to clearly specify the desired image type and purpose, implicitly guiding the model toward retrieval or generation.
    \begin{quote}
        This means the task should not explicitly say whether the model must retrieve or generate the visual, but should instead guide it implicitly through phrases such as ``plot a radar chart'' (generation) or ``include the architecture diagram'' (retrieval).
    \end{quote}
    \item All multimodal elements must serve the core analytical objective closely, and the task should \textbf{add multimodal requirements precisely according to actual research needs}, avoiding any decorative or irrelevant inclusion. For example, use a heatmap to show regional differences in policy effectiveness, a model structure diagram to explain algorithmic innovation, or a time-series line chart to support market growth trends.
    \item When a task requires \textbf{images obtained through web retrieval}, those images must be real and publicly accessible so that the task is genuinely executable.
    \item When a task requires \textbf{visualizations generated by code}, the underlying data sources must be real and publicly available so that the task is genuinely executable.
    \begin{quote}
        Counterexample: the requested image appears reasonable but does not actually exist.

        Task: ``As the head of advanced process R\&D at a semiconductor company, I am leading the push toward mass production at 3\,nm and below, and I now need an analytical report for process roadmap decision-making. The report should: (1) explain the core physical and process reasons why transistor density continues to improve at 3\,nm while yield is under pressure or even declines, and compare FinFET and GAA (especially MBCFET) in terms of gate control capability, parasitic effects, and process complexity; (2) analyze the main defect types in EUV lithography and their formation mechanisms. Please cite data from authoritative 2024--2025 industry reports or papers, and include two schematic figures for explanation: one showing the structural evolution from FinFET to GAA (MBCFET), and another showing \textbf{the formation mechanisms of key EUV lithography defects}, with all key technical parameters and their literature sources labeled in the figure.''
    \end{quote}
\end{itemize}

Note: task descriptions may also randomly include requirements for tables where appropriate, provided that such requirements are genuinely motivated by the task.

\vspace{0.5em}
\textbf{Overall Requirements}
\vspace{0.5em}

We need to construct a total of 120 tasks, with 60 in Chinese and 60 in English. For each language, the tasks should be divided by \textbf{complexity level} into \textbf{low, medium, and high}, with 20 tasks in each level.

Within the 20 tasks at each complexity level, the following \textbf{10 domains} must be covered, strictly following the distribution below:

\begin{itemize}
    \item Technology / Intelligence: 4 tasks (covering frontier areas such as artificial intelligence, computer science, semiconductors, microelectronics, chips, quantum computing, etc.; note: lithium batteries and energy storage technologies belong to ``Environment / Energy'')
    \item Finance / Business: 3 tasks
    \item Health / Medicine: 2 tasks
    \item Law / Policy: 1 task
    \item Environment / Energy: 2 tasks
    \item Academia / Research: 1 task (focusing on frontier progress in foundational disciplines such as mathematics, physics, chemistry, etc.)
    \item Education / Culture: 2 tasks
    \item Literature / Art: 2 tasks
    \item History / Society: 2 tasks
    \item Travel / Entertainment: 1 task
\end{itemize}

\textbf{Localized content that reflects the cultural background of the language is encouraged.}

\vspace{0.5em}
\textbf{Task Examples}
\vspace{0.5em}

\textbf{Data sources:} In an ideal research report, all information acquired through web search should be clearly cited. Therefore, task descriptions may randomly include constraints related to sources, such as ``label image sources,'' etc.

Note: the following examples are constructed for non-specialist researchers and are provided for reference only; they do not necessarily represent the optimal task formulation.

\textbf{1. Low Complexity}

\textbf{About 200 Chinese characters (or about 130 English words); 1--3 multimodal requirements.} The task may simply be described in the form of an outline; bullet points are not mandatory.

\begin{itemize}
    \item As the lead researcher at a prominent AI lab specializing in hybrid intelligence systems, I need to compile a comprehensive report for guiding our 2026 R\&D roadmap on integrating neural-symbolic methods with LLM reasoning. Focus on advancements from 2023-2025, drawing from top conferences like NeurIPS, ICML, and ICLR papers, to outline the evolution of neural-symbolic approaches in enhancing LLM capabilities for logical, mathematical, and causal reasoning. Key elements include: core mechanisms addressing LLM hallucinations and reasoning gaps, comparative performance metrics, and case studies from models like Neuro-Symbolic LLMs or AlphaProof variants. Include a retrieved diagram showing neural-symbolic fusion architectures, and draw a timeline chart depicting major milestones and model releases from 2023-2025, highlighting improvements in reasoning tasks. Provide actionable recommendations for our lab's hybrid model development, emphasizing scalability and ethical considerations. [\textbf{Technology / Intelligence}]

    \item As a supply chain strategy consultant, I am investigating the disruptive impact of Shein and Temu on Amazon's US market dominance. The research must focus on the recent strategic shifts, specifically analyzing the "de minimis" tax loophole challenges and the trend toward local US warehousing by Chinese players. Evaluate Amazon's response through its "Low-Price Store" initiatives and fee restructuring. Use authoritative data from the US International Trade Commission or recent retail logistics white papers. Please plot a bar chart comparing the monthly active user (MAU) growth of Temu, Shein, and Amazon in the US through late 2025. Additionally, include a retrieved schematic diagram illustrating the "Cross-Border E-commerce Logistics Model" (often referred to as the direct-shipping model) to contrast with Amazon's FBA model. [\textbf{Finance / Business}]
\end{itemize}

\textbf{2. Medium Complexity}

\textbf{About 400 Chinese characters (or about 260 English words); 2--4 multimodal requirements.} It is recommended to describe the task in bullet points (3--4 points), with each point containing 1--3 sub-questions.

\begin{itemize}
    \item As a senior policy advisor in the UK Treasury's Tax Policy Division, I need to draft a comprehensive report on the role of inheritance taxes in promoting wealth redistribution, to inform potential reforms in the 2026 Finance Bill, systematically covering these dimensions: (1) Policy Evolution: What key reforms to inheritance tax regimes occurred in major economies like the UK, US, and EU countries from 2023-2025? How do these changes aim to enhance redistribution? Draw a timeline chart to chronicle the sequence of major inheritance tax reforms from 2023 to 2025 across these jurisdictions, incorporating nodes for enactment dates, rate adjustments, and annotations on anticipated redistribution effects to visually trace policy progression and international influences. (2) Redistribution Impacts: Based on 2023-2025 studies, how effective are inheritance taxes at reducing wealth inequality? What evidence shows their progressive nature, and what limitations arise from exemptions like business property relief? Draw a bar chart depicting top marginal inheritance tax rates versus wealth concentration levels in select countries. (3) Evasion and Challenges: What common avoidance strategies, were highlighted in 2024-2025 debates, and in what ways do they erode the intended redistributive objectives of inheritance taxes? Analyze conflicts in data, like underreporting in high-net-worth estates, and evaluate anti-evasion measures in recent legislation. (4) Reform Recommendations: Drawing on cross-country comparisons, what policy options could strengthen inheritance taxes for fairer redistribution? Propose 2-3 feasible UK-specific suggestions, weighing economic impacts and political feasibility against 2025 fiscal constraints. Plot a radar chart covering revenue potential, equity gains, and administrative costs to visualize the trade-offs of these options. [\textbf{Law / Policy}]
    \item As a senior climate resilience specialist at Florida's Department of Environmental Protection, prepare a detailed assessment report on how climate change has amplified hurricane intensities, to support updated coastal adaptation policies for 2026-2030. The report must systematically cover these dimensions: (1) Mechanisms of Intensification: What role do warmer sea surface temperatures play in fueling rapid hurricane intensification, based on 2023-2025 attribution analyses? How do factors like increased atmospheric moisture and reduced wind shear contribute, and what are the observed changes in storm duration? Cite schematic diagrams of hurricane formation processes to illustrate energy transfer from oceans to winds. (2) Empirical Evidence from Recent Storms: How has climate change elevated maximum wind speeds in Atlantic hurricanes during the 2023-2025 period, based on rapid attribution studies and related analyses? What metrics show shifts in Saffir-Simpson categories, and how do regional variations highlight data conflicts? Draw a line chart depicting average wind speed increases over the past three years. (3) Future Projections and Risks: What do CMIP6 models forecast for hurricane intensity under +1.5°C to +2°C warming scenarios by 2030? Discuss how these changes could exacerbate coastal flooding, storm surge heights, and associated economic losses in Florida and the broader Atlantic region, while noting model resolution limitations that affect fine-scale predictions. (4) Policy and Mitigation Recommendations: What adaptive strategies are recommended to mitigate the impacts of intensified hurricanes on Florida's coastal areas, considering the compiled evidence from mechanisms, observations, and projections? Analyze trade-offs in implementation costs versus benefits, drawing on 2024-2025 case studies, and recommend prioritized actions with evidence-based rationale. [\textbf{Environment / Energy}]
\end{itemize}

\textbf{3. High Complexity}

\textbf{About 600 Chinese characters (or about 390 English words); 3--5 multimodal requirements.} The task \textbf{must} be described in bullet points (4--5 points), with each point containing 2--4 sub-questions.

\begin{itemize}
    \item As a Workforce Development Policy Lead, I am evaluating the integration of micro-credentials into the national higher education framework to address the widening technical skills gap. We need to define the future of modular learning in a rapidly automating labor market. Please provide a deep research report on the evolving relationship between micro-credentials and traditional academic degrees: (1) Market Adoption and Employer Valuation: Analyze recent hiring data from platforms like LinkedIn and Indeed regarding the acceptance of industry-recognized micro-credentials (e.g., Google Career Certificates, AWS, Coursera Specializations) versus Bachelor’s degrees for entry-level TMT roles. How has the "Skills-First" hiring trend specifically impacted the salary premium traditionally associated with university degrees? Evaluate the effectiveness of micro-credentials for executive-level upskilling. Plot a bar chart comparing employer "hiring intent" statistics for candidates with "Degree Only" vs. "Degree + Micro-credentials" based on recent workforce surveys. (2) Stackability and Institutional Integration: How are traditional universities (e.g., ASU, Open University) "stacking" short courses into full Master’s or Bachelor’s degrees? Evaluate the progress of the "European Learning Model" (ELM) and digital credential interoperability across borders. How are "Credit Transfer" systems evolving to recognize non-academic learning experiences? Retrieve a diagram illustrating the concept of stacking micro-credentials into full degree programs. (3) Socio-Economic Impact and AI-Driven Reskilling: Analyze the role of micro-credentials in "reskilling" workers displaced by AI and automation. Is there a tangible risk of "credential inflation" where modular learning becomes a mandatory, expensive supplement to traditional degrees? Evaluate the accessibility of these credentials for non-traditional students in rural versus urban areas. Generate a line chart visualizing the growth in enrollment for non-degree vocational credentials versus traditional 4-year degrees over the last five years. (4) Regulatory and Quality Assurance Challenges: Summarize recent legislative efforts in the US and UK to allow federal or student aid for short-term vocational programs (e.g., Short-Term Pell Grant proposals). How are "Quality Labels" for micro-credentials being audited to prevent the rise of "micro-credential mills"? Evaluate the role of industry bodies in certifying the "shelf-life" of fast-depreciating tech skills. Create a table comparing the cost, duration, and median salary outcomes for the top 5 most popular micro-credentials of the past year. [\textbf{Education / Culture}]
    \item As a Cultural Heritage Legal Advisor, I am preparing a legal and strategic brief on the landmark negotiations between the British Museum and the Hellenic Ministry of Culture regarding the Parthenon Marbles. This case is setting a 21st-century precedent for the concept of "Universal Museums". Please provide a deep research report covering: (1) Diplomatic and Legal Framework: Analyze the shift in the British Museum’s stance toward a "cultural exchange" or "long-term loan" rather than permanent transfer. Evaluate the impact of recent UK legislative debates on the British Museum Act 1963 and its ban on "de-accessioning". What are the legal implications of the "ownership vs. custodianship" argument? Incorporate a table comparing the legal arguments for "Retention" (British side) vs. "Restitution" (Greek side) based on international law principles. (2) Public Sentiment and Global Decolonization: How have public opinion polls in the UK and Greece changed regarding the return of the Marbles recently? Analyze the influence of other recent restitutions (e.g., Benin Bronzes from Germany, Nigerian artifacts) on the current negotiation momentum. Evaluate the impact of "Gen Z" sentiment on museum ethics and repatriation policies globally. Create a pie chart visualizing UK public opinion on returning the Parthenon Marbles based on recent YouGov or similar surveys. (3) Technological Solutions: Evaluate the role of high-resolution 3D-scanning and robotic carving (e.g., by the Institute for Digital Archaeology) in creating perfect Pentelic marble replicas for display. Can a "Digital Return" or "Replica Swap" satisfy both cultural preservation and local sovereignty? Analyze the concept of "Digital Ownership" in resolving sharing disputes. Include a retrieved image showing a 3D-scanned reconstruction or a high-fidelity replica of Parthenon sculptures being processed or displayed. (4) The Future of the "Encyclopedic Museum": Analyze the long-term implications for other high-profile disputed artifacts, such as the Rosetta Stone or the Nefertiti Bust. Will a Parthenon settlement trigger a "hollowing out" of Western museums or lead to a new era of collaborative global curation? Evaluate the role of private collections in the restitution landscape. Include a table mapping the top 5 most high-profile disputed artifacts in global museums and their current negotiation status. [\textbf{History / Society}]
\end{itemize}

\textbf{USER PROMPT}
\vspace{0.5em}

Today is \texttt{[DATE]}. Please strictly follow the task design principles described above (\textbf{role-driven}, \textbf{demand-oriented}, \textbf{deep research}, \textbf{frontier-focused}, and \textbf{multimodal integration}), and refer to the format and standards in the task examples. In the \texttt{[DOMAIN]} domain, centered on the topic of \texttt{[TOPIC]}, generate one \texttt{[LANGUAGE--COMPLEXITY]} Deep Research task based on extensive web search of relevant materials.

\end{promptbox}

\subsection{Prompts for Report Preprocessing}
\label{app:preprocessing_prompts}

\begin{promptbox}{Prompt for Reference Entries Extraction}
\textbf{SYSTEM PROMPT}
\vspace{0.5em}

You are an expert academic information extractor. Your task is to rigorously and comprehensively extract all references from the research\_report, and return them in order.

\vspace{0.5em}
\textbf{Extraction Guidelines}
\begin{itemize}
    \item Only extract the references section, not citation facts from the body.
    \item For each reference, return:
    \begin{itemize}
        \item ref\_idx: The citation number N, in order of appearance.
        \item ref\_url: The URL of the reference (if available, otherwise empty string).
    \end{itemize}
\end{itemize}

\textbf{Output Format}
\vspace{0.5em}

Respond with a JSON list. Each item:
\begin{lstlisting}[style=jsonstyle]
{
  "ref_idx": "Citation number",
  "ref_url": "URL of the reference"
}
\end{lstlisting}

\vspace{0.5em}
\textbf{USER PROMPT}
\vspace{0.5em}

Please extract all references from the following research\_report according to the guidelines.

\begin{verbatim}
<research_report>
{report_content}
</research_report>
\end{verbatim}

Remember to output ONLY the JSON list, without any extra explanation or markdown code block.
\end{promptbox}

\begin{promptbox}{Prompt for Fact--Citation Pairs Extraction}
\textbf{SYSTEM PROMPT}
\vspace{0.5em}

You are an expert academic information extractor. Your task is to rigorously and comprehensively extract all citation facts from the research\_report body, in the form of (fact, ref\_idx) pairs.

\vspace{0.5em}
\textbf{Extraction Guidelines}
\vspace{0.5em}

\textbf{1. Scope}  
\begin{itemize}
    \item Only extract citation facts that appear in the \textbf{main text body} and have \textbf{explicit citation markers} (e.g. “[3]”, “[4,5]”).  
    \item Ignore:
    \begin{itemize}
        \item Citations related only to \textbf{figures} (e.g. in figure captions or source notes around figures).
        \item Any \textbf{unmarked} citations (mentions of prior work without explicit numbered markers).
    \end{itemize}
\end{itemize}
\textbf{2. Fact}
\begin{itemize}
    \item Must be complete, understandable sentences or paragraphs containing the citation markers. Expand context if needed for clarity.
    \item When one or more citations are associated with a table:
    \begin{itemize}
        \item If the citations apply to the \textbf{entire table} (e.g., in the table caption or source notes around the table), the fact must contain the \textbf{entire table content}.
        \item If the citations apply to \textbf{only part of the table} (e.g., a specific row, column, or cell), then:
        \begin{itemize}
            \item Create a separate fact for \textbf{each such cited part}.
            \item Each fact must describe \textbf{only that specific part} (the corresponding row/column/cell content), \textbf{not the whole table}.
        \end{itemize}
    \end{itemize}
\end{itemize}
\textbf{3. ref\_idx}
\begin{itemize}
    \item Use the citation numbers exactly as they appear in the text.
    \item If a fact cites multiple references, include all cited numbers in ref\_idxs.
\end{itemize}

\textbf{Output Format}
\vspace{0.5em}

Respond with a JSON list. Each item:
\begin{lstlisting}[style=jsonstyle]
{
  "fact": "Text fragment from the report body containing the citation marker(s)",
  "ref_idxs": ["1", "2"]
}
\end{lstlisting}

\vspace{0.5em}
\textbf{USER PROMPT}
\vspace{0.5em}

Please extract all citation facts from the following research\_report according to the guidelines.

\begin{verbatim}
<research_report>
{report_content}
</research_report>
\end{verbatim}

Remember to output ONLY the JSON list, without any extra explanation or markdown code block.
\end{promptbox}

\begin{promptbox}{Prompt for Figure Elements Extraction}
\textbf{SYSTEM PROMPT}
\vspace{0.5em}

You are an expert academic information extractor. Your task is to rigorously and comprehensively extract all figures from the research\_report, and return them in order.

\vspace{0.5em}
\textbf{Extraction Guidelines}
\begin{itemize}
    \item Only extract figures such as images, charts, graphs, diagrams and etc.
    \item For each figure, return:
    \begin{itemize}
        \item caption: The complete figure caption/title exactly as it appears. (verbatim; do not rewrite or translate).
        \item contents: The file path(s) or URL(s) of the figure.
        \item context: If the figure is explicitly referenced in the main text (e.g., “see Figure 2”, “Figure 2 shows…”), extract the \textbf{sentence(s) or paragraph} that contains the reference and explains it. If there is no explicit reference, extract the \textbf{surrounding text near the figure}, and \textbf{insert the placeholder <figure> on its own line at the exact position where the figure appears}.
        \item ref\_idxs: Include \textbf{only} citation numbers that appear \textbf{within the figure's caption or associated legend} (typically indicating the data source). If none are directly mentioned, return an empty list.
    \end{itemize}
\end{itemize}

\textbf{Output Format}
\vspace{0.5em}

Respond with a JSON list. Each item:
\begin{lstlisting}[style=jsonstyle]
{
  "caption": "Title or caption of the figure",
  "contents": ["Path(s) or URL(s) of the figure"],
  "context": "Contextual text from the report",
  "ref_idxs": ["1", "2"]
}
\end{lstlisting}

\vspace{0.5em}
\textbf{USER PROMPT}
\vspace{0.5em}

Please extract all figures from the following research\_report according to the guidelines.

\begin{verbatim}
<research_report>
{report_content}
</research_report>
\end{verbatim}

Remember to output ONLY the JSON list, without any extra explanation or markdown code block.
\end{promptbox}
\subsection{Prompts for LLM-as-a-Judge Evaluation}
\label{app:judge_prompts}

\begin{promptbox}{Prompt for Citation Support}
\textbf{SYSTEM PROMPT}
\vspace{0.5em}

You are an expert evaluator specializing in fact verification against reference materials. Your task is to rigorously assess whether the statement is supported by the content of its associated references.

\vspace{0.5em}
\textbf{Judgment Criteria}
\begin{itemize}
    \item First, determine whether the references contain any valid information. If they are empty, show “page not found,” or otherwise lack substantive content, the statement should be marked as "unknown".
    \item If the references are valid:
    \begin{itemize}
        \item If a statement's facts or data can be fully found in the references, mark it as "supported" (data can be rounded).
        \item If only part of the facts or data in the statement can be found in the references, mark it as "partially\_supported".
        \item If none of the facts or data in the statement can be found in the references, mark it as "unsupported".
    \end{itemize}
    \item Only judge based on the actual content provided.
\end{itemize}

\textbf{Output Format}
\vspace{0.5em}

Respond with a JSON object. Do not add markdown code blocks (like ```json) around the output.
\begin{lstlisting}[style=jsonstyle]
{
  "result": "supported",
  "justification": "Clearly explain which parts of the statement are supported, partially supported, or unsupported, and cite direct evidence from the references."
}
\end{lstlisting}

\vspace{0.5em}
\textbf{USER PROMPT}
\vspace{0.5em}

Please judge whether the following statement is supported by the references according to the criteria.

\begin{verbatim}
<statement>
{fact}
</statement>

<references_json>
{json.dumps(references_payload, ensure_ascii=False)}
</references_json>
\end{verbatim}

Remember to output ONLY the JSON object.
\end{promptbox}

\begin{promptbox}{Prompt for Instruction Alignment}
\textbf{SYSTEM PROMPT}
\vspace{0.5em}

You are an expert evaluator specializing in auditing multi-modal deep research reports. Your task is to rigorously assess whether a generated research report aligns with a specific checklist derived from a user's research query.

\vspace{0.5em}
\textbf{Goal}
\vspace{0.5em}

Determine if the report \textbf{fully, accurately, and deeply} addresses each item in the checklist.

\vspace{0.5em}
\textbf{Evaluation Scoring Criteria}
\begin{itemize}
    \item \textbf{Score 1 (Pass)}: The report provides a complete, specific, and substantial answer to the checklist item. All requested data, analysis, formats, and timeframes are met.
    \item \textbf{Score 0.5 (Partial)}: The report addresses the checklist item but is missing some required elements, is somewhat vague, only partially meets constraints, or provides incomplete/underspecified evidence. The item is \textit{partially} satisfied.
    \item \textbf{Score 0 (Fail)}: The report fails to provide the required information, is largely vague, misses key constraints, or is missing entirely.
\end{itemize}

\textbf{Important Guidelines}
\vspace{0.5em}

\textbf{1. Explicit Inclusion of Multimodal Elements}
\begin{itemize}
    \item Valid multimodal elements include embedded or clearly shown \textbf{figures}, or a \textbf{concrete, non-empty file path or URL} that clearly points to such an element.  
    \item The following do \textbf{not} qualify as valid multimodal content:  
    \begin{itemize}
        \item Placeholders (e.g., “Figure 1: [to be inserted]”, “image here”).  
        \item Any references without an actual figure or a concrete locator (e.g., “see figure below” but no figure, URL, or file path is provided).  
        \item Any figures shown only as plain text or ASCII-style diagrams. 
    \end{itemize}
\end{itemize}
\textbf{2. Focus on Substantive Data Delivery}
\begin{itemize}
    \item The evaluation must be based strictly on the actual, concrete content delivered in the report; scoring is only permitted when the report explicitly includes substantive content fulfilling the required information.
    \item The following do \textbf{not} qualify as satisfying an item:
    \begin{itemize}
        \item Methodology-only text (plans, approaches, or descriptions of how data might be obtained) without delivered results.
        \item Statements indicating the required information is missing, unavailable, not found, or not provided (e.g., “information not provided,” “no data found”).
    \end{itemize}
\end{itemize}

\textbf{Instructions}
\begin{itemize}
    \item \textbf{Step 1}: For each checklist item, locate the directly relevant section in the "Research Report”.
    \item \textbf{Step 2}: Verify whether the identified content includes all the specific information required by that item.
    \item \textbf{Step 3}: Provide a score from \{0, 0.5, 1\}.
    \item \textbf{Step 4}: Write a clear justification that explicitly states which required elements have been provided, or specifically identifies what is missing or non-compliant, citing direct excerpts from the report as evidence.
\end{itemize}

\textbf{Output Format}
\vspace{0.5em}

Respond with a JSON object mirroring the input structure. Do not add markdown code blocks (like ```json) around the output.
\begin{lstlisting}[style=jsonstyle]
{
  "Section Name": {
    "Checklist Item Text": { "score": 0, "justification": "..." },
    ...
  }
}
\end{lstlisting}

\vspace{0.5em}
\textbf{USER PROMPT}
\vspace{0.5em}

Please evaluate the instruction alignment of the following research report based on the provided checklist.

\begin{verbatim}
<original_query>
{original_query}
</original_query>

<checklist>
{checklist}
</checklist>

<research_report>
{report_content}
</research_report>
\end{verbatim}

Remember to output ONLY the JSON object.
\end{promptbox}

\begin{promptbox}{Prompt for Writing Quality}
\textbf{SYSTEM PROMPT}
\vspace{0.5em}

You are an expert academic editor. Your task is to evaluate the \textbf{Writing Quality} of a model-generated \textbf{research\_report}.

\vspace{0.5em}
\textbf{Goal}
\vspace{0.5em}

Evaluate how well the \textbf{research\_report} functions as academic/professional prose in the style of a research report, focusing on writing quality rather than factual correctness.

\vspace{0.5em}
\textbf{Scoring Dimensions (Each 1–10 points)}
\vspace{0.5em}

\textbf{1. Coherence \& Organization}

\textit{Does the research\_report have a logical, coherent flow at the document and paragraph level, with well-structured sections and smooth transitions?}

Focus: 
\begin{itemize}
    \item Overall structure: clear sections and a reasonable order.  
    \item Paragraph-level coherence: each paragraph has a clear main idea; sentences within a paragraph follow a logical progression.  
    \item Transitions between sections/paragraphs: the reader is guided smoothly from one topic to the next, without abrupt or unexplained jumps.
\end{itemize}

Scoring:
\begin{itemize}
    \item \textbf{9-10 – Excellent coherence and organization}: The report’s structure is clear and well-designed; sections and paragraphs are logically ordered. Transitions are smooth, the progression of ideas is easy to follow, and the overall reading experience feels cohesive from start to finish.
    \item \textbf{7-8 – Good coherence}: The overall structure is clear and sensible. There may be minor rough transitions or slightly uneven sections, but the progression of topics or ideas is consistently understandable and largely well ordered.
    \item \textbf{5-6 – Acceptable but uneven}: A recognizable structure exists, but some paragraphs or sections feel disjointed, misplaced, or weakly connected. Transitions are occasionally abrupt, and the reader sometimes has to work to reconstruct how one part leads to the next.
    \item \textbf{3-4 – Weak coherence}: Organization is confusing or inconsistent. Sections or paragraphs appear in a suboptimal order; topic shifts are abrupt; transitions are often missing. The overall flow is unclear and hard to follow.
    \item \textbf{1-2 – Poorly organized}: The report is largely incoherent in its sequencing. Topics are mixed together without clear structure, transitions are absent, and the reader struggles to understand how the text is organized or how different parts relate to each other.
\end{itemize}

\textbf{2. Clarity \& Readability}

\textit{Is the research\_report written in clear, readable language, avoiding unnecessary complexity and ambiguity?}

Focus:  
\begin{itemize}
    \item Sentence-level clarity: proper grammar, clear syntax, and appropriate vocabulary.  
    \item Avoidance of overly convoluted, ambiguous, or unnecessarily complex sentences.  
    \item Technical terms and specialized concepts are used appropriately and explained or contextualized when necessary.
\end{itemize}

Scoring:
\begin{itemize}
    \item \textbf{9-10 – Very clear and readable}: Sentences are well-constructed and easy to understand. Technical terms are used appropriately and explained when necessary. The text is pleasant to read and accessible to the intended audience.
    \item \textbf{7-8 – Generally clear}: The text is understandable with only minor awkward phrasing or occasional dense passages. Overall readability is good.
    \item \textbf{5-6 – Mixed clarity}: The main ideas can be understood, but there are noticeable awkward sentences, occasional ambiguities, or sections that are harder to parse. The reader needs extra effort at times.
    \item \textbf{3-4 – Hard to read}: Frequent grammar or syntax issues, long and tangled sentences, or confusing wording. Clarity problems significantly hinder comprehension.
    \item \textbf{1-2 – Very unclear}: Serious language problems or extremely convoluted writing make the report difficult to understand or even follow.
\end{itemize}

\textbf{3. Conciseness \& Redundancy}

\textit{Does the research\_report avoid unnecessary repetition and filler, expressing ideas as succinctly as is reasonable for the task defined by the query?}

Focus:  
\begin{itemize}
    \item Presence of repeated arguments, sentences, or paragraphs without added value.  
    \item Overuse of generic phrases, boilerplate, or “throat-clearing” that does not advance the content.  
    \item Whether section length is justified by information content; no obvious padding or “water”.  
\end{itemize}

Scoring:  
\begin{itemize}
    \item \textbf{9-10 – Highly concise}: The report is tight and efficient: almost no redundancy or filler. Each paragraph contributes new information or perspective relevant to the query.
    \item \textbf{7-8 – Mostly concise}: Some minor repetition or slightly wordy passages, but overall the report is reasonably succinct and does not feel bloated.
    \item \textbf{5-6 – Moderate redundancy}: The report contains noticeable repeated ideas or verbose phrasing. While still usable, it could be significantly tightened without loss of content.
    \item \textbf{3-4 – Substantial redundancy}: Many points are repeated, or large portions add little beyond what has already been stated. The report feels padded or overly long relative to its substantive content.
    \item \textbf{1-2 – Highly verbose and repetitive}: Extensive redundancy and filler severely dilute the message. The key ideas are buried under unnecessary text.
\end{itemize}

\textbf{4. Stylistic \& Referencing Consistency}  

\textit{Is the writing style internally consistent in terms of formatting, including its use of tone, terminology, citations, references, and figures?}

Focus:
\begin{itemize}
    \item Tone: the level of formality is stable; the text does not jump needlessly between very informal and highly formal styles.  
    \item Terminology: key terms, labels, and names for the same concept, variable, or method are used consistently.  
    \item Citations: one coherent citation format is followed uniformly.
    \item References: the format of reference list entries is consistent.  
    \item Figures: 
    \begin{itemize}
        \item Numbering format is consistent.  
        \item Caption format is consistent.  
        \item Source attribution format is consistent across all figures.
        \item In-text citation format for figures is consistent.
    \end{itemize}
\end{itemize}

\textbf{Important}: Evaluate only whether these elements are \textbf{internally consistent in formatting} throughout the report. Do NOT evaluate whether they conform to academic standards, whether content quality is appropriate, whether sources are credible, whether information is accurate, whether the number of sources is uniform, or whether source types are uniform. Focus solely on consistency of formatting: Are the same things formatted the same way throughout?

Scoring:
\begin{itemize}
    \item \textbf{9-10 – Highly consistent formatting}: Tone level, terminology spelling/capitalization, citation format, reference list format, and figure formatting are uniform throughout. All formatting choices follow a single consistent pattern.
    \item \textbf{7-8 – Minor formatting inconsistencies}: Overall formatting is consistent, with only occasional small deviations that do not significantly affect the professional appearance.
    \item \textbf{5-6 – Noticeable formatting inconsistencies}: There are several visible inconsistencies in formatting that reduce visual polish but do not prevent understanding.
    \item \textbf{3-4 – Frequent formatting inconsistencies}: The report often switches between different formatting styles for the same elements. Citation formats are mixed, terminology formatting is unstable, and figure formatting varies significantly. This is visually distracting and appears unprofessional.
    \item \textbf{1-2 – Very inconsistent formatting}: Formatting is chaotic with no clear standard. The same terms are formatted differently throughout, citation styles change arbitrarily, reference entries have no consistent format, and figure numbering/captions follow no pattern. The report appears visually disorganized and unprofessional.
\end{itemize}

\textbf{Instructions}
\begin{itemize}
    \item \textbf{Step 1}: Review the provided \textbf{Research\_Report}.  
    \item \textbf{Step 2}: Evaluate the \textbf{Writing Quality} of the Research\_Report across the four dimensions: \textbf{Coherence \& Organization}, \textbf{Clarity \& Readability}, \textbf{Conciseness \& Redundancy}, and \textbf{Stylistic \& Referencing Consistency}.  
    \item \textbf{Step 3}: For each dimension, provide a \textbf{detailed justification} that examines concrete evidence from the report and explains the reasoning leading to the score.
    \item \textbf{Step 4}: Based on the justification, assign a \textbf{1–10 score} for each dimension according to the detailed scoring criteria.
    \item \textbf{Step 5}: If any dimension scores \textbf{6 or lower}, use the \textbf{suggestion} field to propose concrete improvements. If \textbf{all} dimensions score \textbf{7 or higher}, set the \textbf{suggestion} field as an empty string "".
\end{itemize}

\textbf{Output Format}
\vspace{0.5em}

Respond with a JSON object. Do not add markdown code blocks (like ```json) around the output.
\begin{lstlisting}[style=jsonstyle]
{
  "justification": {
    "Scoring Dimension": "specific reasons for the score...",
    ...
  },
  "scores": {
    "Scoring Dimension": 8,
    ...
  },
  "suggestion": "..."
}
\end{lstlisting}

\vspace{0.5em}
\textbf{USER PROMPT}
\vspace{0.5em}

Please evaluate the Writing Quality of the following research report based on the provided rubric.

\begin{verbatim}
<research_report>
{report_content}
</research_report>
\end{verbatim}
Remember to output ONLY the JSON object.

\end{promptbox}

\begin{promptbox}{Prompt for Analytical Depth \& Breadth}
\textbf{SYSTEM PROMPT}
\vspace{0.5em}

You are an expert academic editor. Your task is to evaluate the \textbf{Analytical Depth \& Breadth} of a model-generated \textbf{research\_report}.

You will be given:  
\begin{enumerate}
    \item \textbf{Original\_Query}: The user’s original request or research question.  
    \item \textbf{Research\_Report}: The model-generated report responding to that query.
\end{enumerate}

\textbf{Goal}
\vspace{0.5em}

Assess how deeply and broadly the \textbf{research\_report} thinks about and analyzes the topic, beyond merely restating facts or copying surface-level content.

\vspace{0.5em}
\textbf{Scoring Dimensions (Each 1–10 points)}
\vspace{0.5em}

\textbf{1. Causal Explanatory Reasoning}

\textit{To what extent does the report go beyond description and provide clear, causal or mechanism-oriented explanations for its main claims?}

Focus:  
\begin{itemize}
    \item Does the report move beyond simply stating “what happened” to explain “why” and “how it happens”?  
    \item Are there clear causal or mechanism links (e.g., linking causes, conditions, and outcomes), rather than a loose list of factors?  
    \item Does it specify key causal pathways and the conditions under which they hold, rather than implying causality?
\end{itemize}

Scoring:  
\begin{itemize}
    \item \textbf{9-10 – Excellent causal explanatory reasoning}: The report consistently explains why and how key phenomena occur via explicit causal pathways and mechanisms, clarifying conditions/boundaries for major claims with minimal unexplained jumps.  
    \item \textbf{7-8 – Strong but slightly uneven reasoning}: Most important points include plausible causal/mechanism explanations, though some claims lack clear pathways or boundary conditions.  
    \item \textbf{5-6 – Moderate explanatory reasoning}: Some causal/mechanism explanation is present, but several key claims rely on generic “drivers” language without specifying how the mechanism operates or when it applies.  
    \item \textbf{3-4 – Limited or shallow reasoning}: Explanations are mostly vague (“due to”, “driven by”) with little mechanism detail or conditionality; causal links are often implied rather than articulated.  
    \item \textbf{1-2 – Very poor or purely descriptive}: The report is largely descriptive, with almost no attempts to explain why or how; causal/mechanism thinking is essentially absent.
\end{itemize}

\textbf{2. Analytical Depth \& Development}

\textit{To what extent does the report develop its key points into sustained analysis rather than brief, list-like statements?}

Focus:
\begin{itemize}
    \item Are key claims unpacked with multi-step reasoning rather than asserted in a single step?
    \item Are the most important points developed into coherent, connected paragraphs instead of fragmented one-line bullets?
    \item Does the report prioritize depth on central points, with added length being substantive rather than repetitive?
\end{itemize}

Scoring:
\begin{itemize}
    \item \textbf{9-10 – Excellent development}: Central claims are unpacked with clear multi-step reasoning; key sections show sustained, connected analysis with strong internal structure. Depth is concentrated on the most important points; added length is substantively analytical with minimal filler or repetition.
    \item \textbf{7-8 – Strong development with minor thin spots}: Most key points are developed beyond surface summary and include some multi-step reasoning. Structure is generally coherent, but a few sections remain somewhat compressed, list-like, or stop short of fully unpacking implications/logic.
    \item \textbf{5-6 – Mixed/uneven development}: The report alternates between some developed passages and many thin statements. Several important claims are only partially unpacked, with reasoning that is short, generic, or stops after one step. List-like presentation is common for core points.
    \item \textbf{3-4 – Shallow/list-like development}: Predominantly brief bullets or high-level summaries. Claims are often asserted with little to no reasoning, and important points lack sustained development or connected explanation.
    \item \textbf{1-2 – Very poor development}: Fragmentary and minimally elaborated. The report mostly consists of headings or one-liners with little analytical content and no meaningful development of key points.
\end{itemize}

\textbf{3. Critical Evaluation \& Assumptions}

\textit{To what extent does the report critically examine its assumptions, limitations, and alternative viewpoints instead of treating its own perspective as unquestioned?}

Focus:  
\begin{itemize}
    \item Does the report identify and discuss key assumptions underlying its analysis, models, or narratives?  
    \item Does it acknowledge limitations in data, methods, or perspective, and consider how these might affect the reliability or scope of its conclusions?  
    \item Does it engage with relevant alternative explanations, interpretations, or strategies, rather than presenting a single viewpoint as obviously correct?
\end{itemize}

Scoring:
\begin{itemize}
    \item \textbf{9-10 – Excellent critical evaluation}: The report explicitly surfaces major assumptions and examines their plausibility. It discusses important limitations of data and methods and thoughtfully considers how these constrain the conclusions. Alternative viewpoints or explanations are actively engaged and weighed, showing strong self-critical awareness.  
    \item \textbf{7-8 – Solid but incomplete critical reflection}: The report shows clear critical thinking: it notes some key assumptions, limitations, and alternative perspectives. However, not all important issues are explored in depth, and parts of the analysis still rely on relatively unexamined premises.  
    \item \textbf{5-6 – Moderate critical evaluation}: There is some acknowledgement of assumptions or limitations, and the report may briefly mention alternative views. However, this is sporadic and not integrated into the core reasoning; most arguments still proceed as if their premises are mostly unquestioned.  
    \item \textbf{3-4 – Limited critical reflection}: Assumptions and limitations are largely implicit or mentioned only superficially. Alternative explanations or viewpoints are barely considered, and the report mostly treats its own framing and data as straightforward and unproblematic.  
    \item \textbf{1-2 – Very poor or absent critical evaluation}: The report presents its analysis as if it is fully self-evident, with virtually no discussion of assumptions, limitations, or alternatives. It shows little awareness of potential biases, blind spots, or uncertainties.
\end{itemize}

\textbf{4. Actionable Forward-Looking Insight}

\textit{Does the report translate its analysis into meaningful, forward-looking implications or recommendations that help the reader understand what to do or watch for next?}

Focus:
\begin{itemize}
    \item Does the report derive concrete implications for action, decision-making, strategy, policy, or further work from its analysis?  
    \item Are recommendations or takeaways specific, directional, and conditional (where appropriate), rather than vague or purely formulaic statements?  
    \item Does the report consider future scenarios, risks, opportunities, or monitoring points, instead of stopping at a description of the current state?
\end{itemize}

Scoring:
\begin{itemize}
    \item \textbf{9-10 – Excellent actionable and forward-looking insight}: The report offers clear, specific, and well-justified implications or recommendations that follow from the analysis. It highlights relevant future scenarios, risks, or opportunities and gives the reader a strong sense of “what to do or pay attention to next.”  
    \item \textbf{7-8 – Strong but somewhat limited insight}: There are meaningful, reasonably specific implications or recommendations, and some forward-looking reflection. However, coverage is not fully systematic: certain conclusions are left without actionable follow-through or future-oriented consideration.  
    \item \textbf{5-6 – Moderate actionable insight}: The report contains some useful takeaways, but they are either unevenly developed, somewhat generic, or only loosely tied to the preceding analysis. Forward-looking elements are present but not deeply integrated.  
    \item \textbf{3-4 – Limited or generic implications}: Implications and recommendations, if present, are mostly high-level, vague, or boilerplate with little specificity or prioritization. There is little sense of future scenarios or concrete next steps.  
    \item \textbf{1-2 – Very poor or absent forward-looking content}: The report mostly describes or analyzes the current situation with almost no attempt to derive practical implications or future-oriented insights.
\end{itemize}

\textbf{5. Thematic Breadth \& Coverage}

\textit{How comprehensively does the report cover the key aspects of its topic, showing a sufficiently broad and well-chosen thematic scope without obvious blind spots?}

Focus:  
\begin{itemize}
    \item Does the report address the main dimensions that are clearly relevant to the core topic, rather than focusing narrowly on a single aspect?  
    \item Does it reasonably extend beyond only the explicitly mentioned points in the original\_query to include other clearly important, closely related angles?  
    \item While remaining focused, does the report avoid obvious gaps where a major, relevant facet of the topic is missing or severely underdeveloped?
\end{itemize}

Scoring:  
\begin{itemize}
    \item \textbf{9-10 – Excellent thematic breadth and coverage}: The report covers the topic in a well-balanced, comprehensive way, addressing the major relevant dimensions with appropriate depth. It not only responds to the explicitly stated points in the original\_query but also proactively includes other clearly important, closely related angles. This extension beyond the original\_query feels natural and necessary rather than off-topic. There are no clear blind spots.  
    \item \textbf{7-8 – Strong but slightly uneven coverage}: Most important aspects of the topic are covered, and the scope feels suitably broad. The report does move beyond the exact wording of the original\_query to touch on some additional, relevant facets, but one or two significant dimensions are treated briefly or somewhat underdeveloped. The overall breadth is good, with only minor gaps.  
    \item \textbf{5-6 – Moderate thematic breadth}: The report addresses several key facets of the topic, but coverage is selective or uneven. It may mostly stick to the points explicitly mentioned in the original\_query, with only limited extension to other relevant angles, or it may treat some dimensions in depth while leaving others only hinted at. Overall breadth is acceptable but leaves noticeable room for a more complete picture.  
    \item \textbf{3-4 – Narrow or imbalanced coverage}: The report focuses heavily on a limited subset of relevant aspects and largely stays within the most obvious or explicitly stated parts of the original\_query. It rarely brings in additional, closely related angles that would help complete the thematic picture. As a result, the scope feels constrained or skewed, and important dimensions of the topic are missing or severely underdeveloped.  
    \item \textbf{1-2 – Very poor or severely narrow coverage}: The report’s treatment of the topic is extremely narrow, fragmentary, or misaligned. It may restrict itself almost entirely to a small portion of what the original\_query mentions, ignoring other clearly relevant aspects, and it does not attempt to go beyond the original\_query’s explicit wording in any meaningful way. The reader is left without a coherent overall understanding of the subject.
\end{itemize}

\textbf{Instructions}
\begin{itemize}
    \item \textbf{Step 1}: Review the provided \textbf{Original\_Query} and \textbf{Research\_Report}.  
    \item \textbf{Step 2}: Evaluate the \textbf{Analytical Depth \& Breadth} of the Research\_Report across the five dimensions: \textbf{Causal Explanatory Reasoning}, \textbf{Analytical Depth \& Development}, \textbf{Critical Evaluation \& Assumptions}, \textbf{Actionable Forward-Looking Insight}, and \textbf{Thematic Breadth \& Coverage}.  
    \item \textbf{Step 3}: For each dimension, provide a \textbf{detailed justification} that examines concrete evidence from the report and explains the reasoning leading to the score.
    \item \textbf{Step 4}: Based on the justification, assign a \textbf{1–10 score} for each dimension according to the detailed scoring criteria.
    \item \textbf{Step 5}: If any dimension scores \textbf{6 or lower}, use the \textbf{suggestion} field to propose concrete improvements. If \textbf{all} dimensions score \textbf{7 or higher}, set the \textbf{suggestion} field as an empty string "".
\end{itemize}

\textbf{Output Format}
\vspace{0.5em}

Respond with a JSON object. Do not add markdown code blocks (like ```json) around the output.
\begin{lstlisting}[style=jsonstyle]
{
  "justification": {
    "Scoring Dimension": "specific reasons for the score...",
    ...
  },
  "scores": {
    "Scoring Dimension": 8,
    ...
  },
  "suggestion": "..."
}
\end{lstlisting}

\vspace{0.5em}
\textbf{USER PROMPT}
\vspace{0.5em}

Please evaluate the Analytical Depth \& Breadth of the following research report based on the provided rubric.

\begin{verbatim}
<original_query>
{original_query}
</original_query>

<research_report>
{report_content}
</research_report>
\end{verbatim}
Remember to output ONLY the JSON object.

\end{promptbox}

\begin{promptbox}{Prompt for Factual \& Logical Consistency}
\textbf{SYSTEM PROMPT}
\vspace{0.5em}

You are an expert academic editor. Your task is to evaluate the \textbf{Factual \& Logical Consistency} of a model-generated \textbf{research\_report}.

\vspace{0.5em}
\textbf{Goal}
\vspace{0.5em}

Determine whether the report \textbf{contains factual or logical contradictions}, and count the total number of distinct contradictions.

\vspace{0.5em}
\textbf{Out Of Scope}
\begin{itemize}
    \item \textbf{Citation and reference issues}: Do NOT evaluate any problems related to citations or references. This includes inconsistencies in attribution, source identification, or provenance of methods/concepts/data that arise from how citation markers are used or organized. \textbf{If a contradiction exists only because of citation or reference-list issues, it is OUT OF SCOPE.} Only identify contradictions that would still exist if all citation markers and reference entries were removed from the report.
    \item \textbf{External factual accuracy}: Do NOT evaluate whether the report’s statements are true in the real world; only check whether the report contradicts itself internally.  
    \item \textbf{Terminology consistency}: Do NOT evaluate whether key terms, labels, or names for the same concept, variable, or method are used consistently throughout the report; only when such terminological differences lead to a direct contradiction about the same underlying object or result should it be counted as an issue.
\end{itemize}

\textbf{Contradiction Definition}
\vspace{0.5em}

An \textbf{contradiction (issue)} exists when two or more statements in the report, taken at face value within the report’s own context, \textbf{cannot all be true at the same time}, and the report does not provide a clear explanation, condition, or update that resolves this conflict.

You must consider both:

\textbf{1. Factual contradictions}

Examples include, but are not limited to:
\begin{itemize}
    \item The same quantity (e.g., sample size, number of companies, growth rate, percentage, etc.) is given conflicting values with no explanation (e.g., no mention of different subsamples, filtering, or time points).  
    \item The same time period, region/market, or population/scope is described differently in different places for what is clearly the same analysis or dataset.  
    \item The same dataset’s source or status (complete vs missing/bias, etc.) is described in mutually conflicting ways for the same stage of analysis.  
    \item The same result (same metric, same time period) is described with opposite trends or directions in different parts of the text (e.g., “increasing” vs “decreasing”) without clarification.
\end{itemize}

\textbf{2. Logical contradictions}

Examples include, but are not limited to:
\begin{itemize}
    \item The report states an explicit premise, assumption, or applicability condition, and later reasoning or conclusions clearly violate that premise under the same context.  
    \item The report gives a certain judgment/evaluation of a model, strategy, or view in one place and later uses an opposite characterization of the same item in the same context, without new evidence or conditions.  
    \item The conclusions or recommendations directly state the opposite of the report’s own presented results (e.g., results show “unlikely to succeed,” conclusion claims “almost certain to succeed”).  
    \item The report explicitly labels its results as non-generalizable or restricted, but later presents them as broadly applicable without any additional justification.
\end{itemize}

\textbf{Counting Rules}
\begin{itemize}
    \item Each distinct contradiction (factual or logical) counts as \textbf{one} issue.  
    \item Repeated restatements of the same underlying contradiction count as \textbf{one} issue.  
    \item If multiple statements all conflict on the same fact or conclusion, treat that as \textbf{one} issue.  
\end{itemize}

\textbf{Instructions}
\begin{itemize}
    \item \textbf{Step 1}: Read the entire \textbf{Research\_Report}.  
    \item \textbf{Step 2}: Identify all contradictions where the report’s own statements cannot all be true at the same time and no clear explanation is given.  
    \item \textbf{Step 3}: For each identified issue:  
    \begin{itemize}
        \item Provide a brief description of the contradiction.  
        \item Provide textual evidence: quote or closely paraphrase the relevant conflicting statements with location references (section/paragraph labels if available, or your own brief locator such as “Introduction, paragraph 2”).  
    \end{itemize}
\end{itemize}

\textbf{Remember}
\vspace{0.5em}

Be as strict as possible in identifying contradictions. Scrutinize the report thoroughly to find any potential logical or factual inconsistencies.

\vspace{0.5em}
\textbf{Output Format}
\vspace{0.5em}

Please respond with a JSON object. Do not add markdown code blocks (like ```json) around the output.
\begin{lstlisting}[style=jsonstyle]
{
  "total_issues": 4,
  "issues": [
    {  
      "id": 1,
      "description": "brief summary of the contradiction",
      "evidence": {
        "section/paragraph reference": "excerpt or close paraphrase",
        ...
      }
    },
    ...
  ]
}
\end{lstlisting}

\vspace{0.5em}
\textbf{USER PROMPT}
\vspace{0.5em}

Please evaluate the Factual \& Logical Consistency of the following research report based on the provided rubric.

\begin{verbatim}
<research_report>
{report_content}
</research_report>
\end{verbatim}

Remember to output ONLY the JSON object.

\end{promptbox}

\begin{promptbox}{Prompt for Chart Quality}
\textbf{SYSTEM PROMPT}
\vspace{0.5em}

You are an expert Data Visualization Quality Assurance Specialist. Your task is to rigorously audit the visual quality of charts in a research report using a strict \textbf{Binary Checklist}.

\vspace{0.5em}
\textbf{Goal}
\vspace{0.5em}

Determine if the visual chart \textbf{fully, accurately, and clearly} meets the visual quality checklist items regarding layout, readability, and conciseness.

\vspace{0.5em}
\textbf{Evaluation Scoring Criteria}
\begin{itemize}
    \item \textbf{Score 1 (Pass)}: The chart is visually flawless regarding the specific criterion. No issues are visible.
    \item \textbf{Score 0 (Fail)}: The chart fails the criterion. Any visible defect (e.g., text overlap, unreadable font, broken encoding) results in a score of 0.
\end{itemize}

\textbf{Checklist Items To Evaluate}
\vspace{0.5em}

\textbf{A. Layout \& Structure}
\begin{itemize}
    \item \textbf{No Overlap}: All text elements (titles, legends, axis labels, tick labels, annotations, data labels) and graphic elements do not overlap with each other or with the core data visualization components.
    \item \textbf{No Misalignment}: Elements that should align (axis lines, tick marks with tick labels, legend markers with legend text, subplot panels in a grid) are consistently aligned, without clear, unintended offset or irregular positioning.
    \item \textbf{Complete Visibility}: All chart content is fully visible within the canvas boundaries, with no elements cut off, clipped, or extending beyond the viewable area.
    \item \textbf{Balanced Spatial Composition}: The chart's overall composition is balanced; the plot area, titles, legends, and annotations occupy appropriate proportions of the canvas; whitespace is distributed reasonably, appearing neither cramped nor empty.
\end{itemize}

\textbf{B. Readability}  
\begin{itemize}
    \item \textbf{Appropriate Text-to-Chart Proportion}: Text elements (titles, legends, axis labels, tick labels, data labels) are proportionate to the chart’s visual elements: text is neither dominating the plot area nor so small relative to marks/axes that it harms readability.
    \item \textbf{Readable Contrast \& Distinguishability}: Text is clearly readable against the background, and different series/categories are visually distinguishable. Fail if low contrast or overly similar colors make text or categories hard to read/identify.
    \item \textbf{Correct Text Rendering}: All characters, symbols, and numbers are displayed correctly, without encoding errors, missing glyphs, or rendering issues.
\end{itemize}

\textbf{C. Conciseness}
\begin{itemize}
    \item \textbf{No Excessive Decoration}: The chart does not contain distracting or excessive decorative elements; all visual effects are restrained and purposeful, enhancing rather than obscuring data readability.
    \item \textbf{No Information Overload}: The number of visual elements in the chart is not excessive; the chart avoids being too dense or overcrowded, which would cause difficulty in interpretation.
    \item \textbf{No Redundant Labels}: Labels are used efficiently; the same information is not repeated or presented redundantly in ways that create visual clutter.
\end{itemize}

\textbf{Output Format}
\vspace{0.5em}

Respond with a JSON object. Do not add markdown code blocks (like ```json) around the output.
\begin{lstlisting}[style=jsonstyle]
{
  "scores": {
    "overlap": 0,
    "misalignment": 1,
    "visibility": 1,
    "spatial_composition": 1,
    "text_size": 0,
    "contrast": 1,
    "text_rendering": 1,
    "excessive_decoration": 1,
    "information_overload": 1,
    "redundant_labels": 1
  },
  "justification": {
    "overlap": "Specific description of overlap issue...",
    "text_size": "Specific description of font size issue..."
  },
  "suggestion": "Actionable advice to fix the issues..."
}
\end{lstlisting}

\vspace{0.5em}
\textbf{USER PROMPT}
\vspace{0.5em}

Please evaluate the visual quality of the following chart based on the provided checklist. Remember to output ONLY the JSON object.

\textit{The user input also includes the chart itself.
}
\end{promptbox}

\begin{promptbox}{Prompt for Multimodal Composition}
\textbf{SYSTEM PROMPT}
\vspace{0.5em}

You are an expert evaluator specializing in auditing multi-modal research reports. Your task is to assess the \textbf{Multimodal Composition} of a model-generated \textbf{research\_report}.

\vspace{0.5em}
\textbf{Goal}
\vspace{0.5em}

Evaluate how effectively the \textbf{research\_report} organizes and incorporates multimodal elements in its overall document design, with particular attention to the \textbf{layout, quantity, variety, and richness} of figures. Focus on document-level multimodal composition, rather than the local relevance or factual correctness of individual figures.

\vspace{0.5em}
\textbf{Scoring Dimensions (Each 1–10 points)}
\vspace{0.5em}

\textbf{1. Multimodal Layout \& Quantity Appropriateness}

\textit{Are multimodal elements placed, distributed, and used in an appropriate amount across the report’s stages, supporting a smooth reading flow without obvious gaps or overload?}

Focus:  
\begin{itemize}
    \item Figures are placed near the sections where they are discussed, not arbitrarily clustered or detached.  
    \item The overall visual distribution supports a natural reading rhythm, avoiding both long stretches with no visuals despite complex content and sudden dense clusters that disrupt the flow.  
    \item Important complex content (e.g., structures, workflows, data comparisons) has at least some visual support, rather than relying solely on dense text.  
    \item The total number of figures feels proportionate to the length and complexity of the report (neither too few nor excessive).  
    \item Multimodal elements are reasonably spread across key parts, rather than confined to only one section while other equally complex sections have none.
\end{itemize}

Scoring:
\begin{itemize}
    \item \textbf{9-10 – Excellent layout and quantity appropriateness}: Multimodal elements are consistently placed near relevant text and are well distributed across the report’s stages. Coverage is well targeted (no clear gaps), and the quantity feels proportionate with no obvious redundancy. Overall, the layout and amount clearly support a smooth, coherent reading experience.
    \item \textbf{7-8 – Good layout and quantity with minor issues}: Placement and distribution are mostly sensible with occasional issues (a figure could be closer to its discussion, slight clustering, or a small coverage gap). Quantity is generally appropriate, with only minor under/over-use that does not strongly affect readability.
    \item \textbf{5-6 – Acceptable but uneven}: The report remains readable, but layout and/or quantity feel insufficiently planned. There may be noticeable stretches lacking visuals despite complexity, awkward groupings, uneven coverage across sections, or a quantity that is somewhat misaligned (a bit sparse or somewhat cluttered).
    \item \textbf{3-4 – Weak layout/quantity choices}: Figures are frequently detached from relevant text or awkwardly grouped; some key sections lack needed visuals while others feel cluttered. Coverage and quantity choices noticeably hinder reading flow and understanding.
    \item \textbf{1-2 – Very poor layout and quantity appropriateness}: Figures (if any) are placed with little regard to structure and discussion points, and the report is either severely under-illustrated despite complexity or overloaded with visuals. The multimodal design seriously disrupts or confuses the reading experience.
\end{itemize}

\textbf{2. Multimodal Variety \& Richness}

\textit{Does the report demonstrate appropriate variety and richness in its multimodal elements based on the report's topic and nature?}

Focus:
\begin{itemize}
    \item \textbf{Source-type variety}:
    \begin{itemize}
        \item \textbf{Retrieved external images}: e.g., architecture diagrams, system schematics, flowcharts, sequence diagrams, environment photos, illustrations, etc.
        \item \textbf{Code-generated charts}: e.g., line charts, bar charts, scatter plots, heatmaps, radar charts, box plots, etc.
    \end{itemize}
    \item \textbf{Intra-type variety}: Within each source type present, the report uses multiple appropriate sub-types.
    \begin{itemize}
        \item \textbf{For charts}: Different chart types count as different sub-types. (e.g., line chart, bar chart, scatter plot are 3 distinct sub-types).
        \item \textbf{For images}: Different purposes count as different sub-types (e.g., offshore wind power architecture diagram, reinforcement learning algorithm architecture diagram are 2 distinct sub-types).
    \end{itemize}
    \item \textbf{Topic-appropriate multimodal strategy}: The choice and balance of multimodal elements should align with the report's subject matter.
    \begin{itemize}
        \item \textbf{Data/analysis-heavy topics} (e.g., financial analysis, statistical studies, performance benchmarks): Naturally emphasize code-generated charts with rich variety.
        \item \textbf{Technical/architectural topics} (e.g., system design, software architecture, infrastructure): Naturally emphasize retrieved images with rich variety.
        \item \textbf{Balanced topics} (e.g., comprehensive surveys, product analyses): Should include both types with reasonable variety in each.
    \end{itemize}
\end{itemize}

Hard Constraints:
\begin{itemize}
    \item You MUST compute these intermediate variables:
    \begin{itemize}
        \item topic\_category: string in \{"data\_heavy", "technical\_architectural", "balanced"\}
        \item source\_types\_present: integer in \{0,1,2\}
        \item image\_subtypes\_count: integer (0 if no valid images)
        \item chart\_subtypes\_count: integer (0 if no valid charts)
        \item total\_subtypes: integer (image\_subtypes\_count + chart\_subtypes\_count)
        \item dominant\_subtypes\_count: integer (chart\_subtypes\_count for data\_heavy, image\_subtypes\_count for technical\_architectural)
    \end{itemize}

    \item ABSOLUTE SCORING RULES - NO EXCEPTIONS:
    \begin{itemize}
   \item  Once the numerical thresholds are met, the score range is LOCKED and cannot be changed.
   \item  You MUST NOT apply any additional subjective criteria beyond the defined rules, including but not limited to:
    \begin{itemize}
        \item Subjective quality judgments (e.g., "not especially diverse", "basic/common types")
        \item Suggestions for additional content (e.g., "could include more advanced visualizations", "lacks X type")
        \item Personal preferences about specific chart or image types
    \end{itemize}
   \item  The score range is determined EXCLUSIVELY by these objective factors:
    \begin{itemize}
        \item The COUNT of subtypes (as computed above)
        \item Whether both source types are present (for balanced topics only)
        \item The topic category classification
    \end{itemize}
   \item  Within the determined score range (e.g., 9-10 or 7-8), you may select the specific score based on the minor execution quality differences. BUT you CANNOT move to a different score range under any circumstances.
   \end{itemize}

    \item You MUST determine the score by STRICTLY following the mapping below. The score ranges are \textbf{hard boundaries} that you CANNOT violate.
\end{itemize}

Score Mapping (EVALUATE IN ORDER, STOP AT FIRST MATCH):

\begin{itemize}

    \item If source\_types\_present == 0:
    
    \quad  → score MUST be 1 or 2. STOP.
    
    \item If topic\_category == "balanced":
    \begin{itemize}
        \item If source\_types\_present == 2:
        \begin{itemize}
            \item If image\_subtypes\_count >= 3 AND chart\_subtypes\_count >= 3:

            \quad → score MUST be 9 or 10. STOP.

            \item Else if image\_subtypes\_count >= 2 AND chart\_subtypes\_count >= 2:

            \quad → score MUST be 7 or 8. STOP.

            \item Else:

            \quad → score MUST be 5 or 6. STOP.
        \end{itemize}
        \item If source\_types\_present == 1:
        \begin{itemize}
            \item If max(image\_subtypes\_count, chart\_subtypes\_count) >= 3:

            \quad → score MUST be 5 or 6. STOP.
            \item Else:

            \quad → score MUST be 3 or 4. STOP.
        \end{itemize}
    \end{itemize}

    \item If topic\_category in \{"data\_heavy", "technical\_architectural"\}:
    \begin{itemize}
        \item If dominant\_subtypes\_count >= 4:

        \quad → score MUST be 9 or 10. STOP.
        \item Else if dominant\_subtypes\_count >= 3:

        \quad → score MUST be 7 or 8. STOP.
        \item Else if total\_subtypes >= 2:

        \quad → score MUST be 5 or 6. STOP.

        \item Else:
        
        \quad → score MUST be 3 or 4. STOP.
        
    \end{itemize}
\end{itemize}

\vspace{0.5em}
\textbf{Important Guidelines}
\begin{itemize}
    \item Only \textbf{actual, present multimodal elements} are considered valid for this evaluation.  
    \item Valid elements include embedded or clearly shown \textbf{figures}, or a \textbf{concrete, non-empty file path or URL} that clearly points to such an element.  
    \item The following do \textbf{not} count as valid multimodal content:  
    \begin{itemize}
        \item Placeholders (e.g., “Figure 1: [to be inserted]”, “image here”).  
        \item Any references without an actual figure or a concrete locator (e.g., “see figure below” but no figure, URL, or file path is provided).  
        \item Any figures shown only as plain text or ASCII-style diagrams. 
    \end{itemize}
    \item If the report contains \textbf{no valid multimodal elements} under these rules, reflect this in your scores (e.g., very weak layout, quantity and variety) and explicitly mention the absence of real multimodal content in your justification and suggestions.
\end{itemize}

\textbf{Instructions}
\begin{itemize}
    \item \textbf{Step 1}: Review the provided \textbf{Research\_Report}.  
    \item \textbf{Step 2}: Evaluate the \textbf{Multimodal Composition} of the Research\_Report across the two dimensions: \textbf{Multimodal Layout \& Quantity Appropriateness}, and \textbf{Multimodal Variety \& Richness}.  
    \item \textbf{Step 3}: For each dimension, provide a \textbf{detailed justification} that examines concrete evidence from the report and explains the reasoning leading to the score.
    \item \textbf{Step 4}: Based on the justification, assign a \textbf{1–10 score} for each dimension according to the detailed scoring criteria.
    \item \textbf{Step 5}: If any dimension scores \textbf{6 or lower}, use the \textbf{suggestion} field to propose concrete improvements. If \textbf{all} dimensions score \textbf{7 or higher}, set the \textbf{suggestion} field as an empty string "".
\end{itemize}

\textbf{Output Format}
\vspace{0.5em}

Respond with a JSON object. Do not add markdown code blocks (like ```json) around the output.
\begin{lstlisting}[style=jsonstyle]
{
  "justification": {
    "Scoring Dimension": "specific reasons for the score...",
    ...
  },
  "scores": {
    "Scoring Dimension": 8,
    ...
  },
  "suggestion": "..."
}
\end{lstlisting}

\vspace{0.5em}
\textbf{USER PROMPT}
\vspace{0.5em}

Please evaluate the Multimodal Composition of the following research report based on the provided rubric.

\begin{verbatim}
<research_report>
{report_content}
</research_report>
\end{verbatim}

Remember to output ONLY the JSON object.

\end{promptbox}

\begin{promptbox}{Prompt for Figure Caption Quality}
\textbf{SYSTEM PROMPT}
\vspace{0.5em}

You are an expert academic editor and visual analyst. Your task is to evaluate the \textbf{Quality of the Figure Caption} in a research report.

You may be provided with:
\begin{enumerate}
    \item \textbf{Figure}: The visual content (such as a chart, graph, photo, diagram, etc.).
    \item \textbf{Caption}: The full text associated with the figure.
\end{enumerate}

\textbf{Scoring Dimensions (Each 1–10 points)}
\vspace{0.5em}

\textbf{1. Visual Accuracy}

\textit{Does the caption correctly identify and describe what is actually shown?}

Focus:
\begin{itemize}
    \item Alignment between caption and figure.
    \item Correct identification of subject, variables, and visible relationships.
\end{itemize}

Scoring:
\begin{itemize}
    \item \textbf{9-10 – Perfect match}: The caption correctly identifies what the figure is (e.g., chart/diagram/photo), names the key entities/components/variables, and accurately describes the visible content. All claims are consistent with the figure.
    \item \textbf{7-8 – High accuracy}: The main identification and description are correct; minor visible elements are omitted or slightly under-specified, but nothing is wrong or misleading.
    \item \textbf{5-6 – Partial match}: The caption is too generic or incomplete: it roughly matches the topic, but misses key visible elements needed to understand what is shown.
    \item \textbf{3-4 – Low accuracy}: The caption refers to related but materially different content (e.g., wrong type of figure, wrong components/variables, or mischaracterized relationship/comparison).
    \item \textbf{1-2 – Contradiction}: The caption describes content not present in the figure, or asserts relationships/structures/process steps that are clearly contradicted by what is shown.
\end{itemize}

\textbf{2. Minimum Necessary Information}

\textit{Using only the figure and this caption, can a reader correctly interpret what is shown? Does the caption provide the \textbf{minimum necessary} information given the figure’s complexity (no less, no more)?}

Focus:
\begin{itemize}
    \item If a reader can understand the figure’s main point using \textbf{the figure itself + this caption} (without the main text), the caption is sufficient.
    \item Provide only the \textbf{minimal guidance needed to read this figure}. \textbf{Do NOT require} the caption to restate or list details already visible in the figure.
    \item A caption can be short and still score high if the figure is self-explanatory. A longer caption scores high only if the added details are necessary to remove ambiguity for this specific figure.
\end{itemize}

Scoring:
\begin{itemize}
    \item \textbf{9–10 – Minimally sufficient and perfectly calibrated}: The caption provides just enough information for correct interpretation (given the figure), with little or no ambiguity, and without redundant listing or filler content.
    \item \textbf{7–8 – Sufficient with minor inefficiency}: Interpretation is possible without the main text; one small clarification is missing \textbf{or} there is a small amount of unnecessary detail, but overall still clear.
    \item \textbf{5–6 – Borderline}: The caption misses 1–2 key clarifications needed to interpret the figure \textbf{or} includes noticeable non-essential detail that does not improve interpretability; a reader may need the main text.
    \item \textbf{3–4 – Insufficient or bloated}: Multiple necessary clarifications are missing \textbf{and/or} the caption is mostly verbose filler, leading to confusion or ambiguity.
    \item \textbf{1–2 – Not usable}: The caption is too vague/empty to interpret the figure, or is largely irrelevant to what is shown, making independent interpretation impossible.
\end{itemize}

\textbf{3. Clarity \& Readability}

\textit{Is the caption written in clear and accessible language that is easy to understand?}

Focus:
\begin{itemize}
    \item Clear sentence structure and word choice.
    \item Avoids unnecessary jargon or defines technical terms when necessary.
    \item Easy to parse and understand on first reading.
\end{itemize}

Scoring:
\begin{itemize}
    \item \textbf{9-10 – Highly clear}: The caption uses straightforward, direct language that is immediately understandable. Technical terms are either avoided or clearly defined. No ambiguous phrasing.
    \item \textbf{7-8 – Clear}: Generally easy to understand with minor issues (e.g., one slightly complex sentence or undefined term that doesn't significantly hinder comprehension).
    \item \textbf{5-6 – Moderately clear}: Contains some unclear phrasing, unnecessary complexity, or undefined jargon that requires re-reading or may confuse readers.
    \item \textbf{3-4 – Unclear}: Multiple confusing sentences, heavy jargon without explanation, or overly complex language that obscures the main point.
    \item \textbf{1-2 – Incomprehensible}: The caption is so poorly written, convoluted, or jargon-heavy that readers would struggle to understand what is being described.
\end{itemize}

\textbf{Important Guidelines}
\begin{itemize}
    \item You must primarily compare the \textbf{Caption} against the \textbf{Visual Evidence} in the \textbf{Figure}.  
    \item Do \textbf{not} invent or assume facts that are not supported by the figure.
\end{itemize}

\textbf{Instructions}
\begin{itemize}
    \item \textbf{Step 1}: Review the provided \textbf{Figure} and \textbf{Caption}.
    \item \textbf{Step 2}: Evaluate the \textbf{Quality of the Figure Caption} across the three dimensions: \textbf{Visual Accuracy}, \textbf{Minimum Necessary Information}, and \textbf{Clarity \& Readability}.
    \item \textbf{Step 3}: For each dimension, provide a \textbf{detailed justification} that examines concrete evidence from the figure and explains the reasoning leading to the score.
    \item \textbf{Step 4}: Based on the justification, assign a \textbf{1–10 score} for each dimension according to the detailed scoring criteria.
    \item \textbf{Step 5}: If any dimension scores \textbf{6 or lower}, provide a \textbf{revised caption} in the \textbf{suggestion} field, using only information supported by the figure. If \textbf{all} dimensions score \textbf{7 or higher}, set the \textbf{suggestion} field as an empty string "".
\end{itemize}

\textbf{Output Format}
\vspace{0.5em}

Respond with a JSON object. Do not add markdown code blocks (like ```json) around the output.
\begin{lstlisting}[style=jsonstyle]
{
  "justification": {
    "Scoring Dimension": "specific reasons for the score...",
    ...
  },
  "scores": {
    "Scoring Dimension": 8,
    ...
  },
  "suggestion": "..."
}
\end{lstlisting}

\vspace{0.5em}
\textbf{USER PROMPT}
\vspace{0.5em}

Please evaluate the Quality of the Figure Caption based on the provided rubric.

\begin{verbatim}
<caption>
{caption}
</caption>
\end{verbatim}
Remember to output ONLY the JSON object.

\textit{The user input also includes the figure itself.
}
\end{promptbox}

\begin{promptbox}{Prompt for Figure--Context Integration}
\textbf{SYSTEM PROMPT}
\vspace{0.5em}

You are an expert academic editor and visual analyst. Your task is to evaluate how well a \textbf{Figure} is integrated with its surrounding \textbf{Context} in a research report.

You will be given:
\begin{enumerate}
    \item \textbf{Figure}: The visual content (such as a chart, graph, photo, diagram, etc.).
    \item \textbf{Context}: The paragraph(s) in the report that reference or surround this figure. The placeholder <figure> indicates where the figure appears in the text.
\end{enumerate}

\textbf{Goal}
\vspace{0.5em}

Judge the \textbf{relationship between the Figure and the Context}.

\vspace{0.5em}
\textbf{Scoring Dimensions (Each 1–10 points)}
\vspace{0.5em}

\textbf{1. Contextual Relevance}

\textit{How well does the figure's content match the specific topic and claims in the context?}

Focus:
\begin{itemize}
    \item \textbf{Semantic alignment} between what the figure shows and what the context discusses.
    \item Whether the figure depicts the same topic, variables, phenomena, or cases mentioned in the surrounding text.
\end{itemize}

Scoring:
\begin{itemize}
    \item \textbf{9-10 – Highly Relevant}: The figure directly depicts or represents the specific subject matter discussed in the immediate context. A reader can clearly see why this particular figure was chosen for this passage.
    \item \textbf{7-8 – Relevant}: The figure clearly depicts the specific subtopic or subject in this passage, though it may not capture every detail mentioned in the text. The figure choice is appropriate for this context.
    \item \textbf{5-6 – Moderately Relevant}: The figure belongs to the same broader theme or research area and is not out of place, but the connection between what the figure shows and what this passage specifically discusses is loose.
    \item \textbf{3-4 – Weak Relevance}: The connection to the context is vague or forced. It only shares a very general domain, without depicting what this passage is specifically about.
    \item \textbf{1-2 – Irrelevant}: The figure does not depict anything meaningfully related to the local context; a typical reader would be confused why this particular figure is placed here.
\end{itemize}

\textbf{2. Narrative Coherence}

\textit{How smoothly is the figure integrated into the flow of the text?}

Focus:
\begin{itemize}
    \item Whether the context \textbf{explicitly or clearly implicitly} points the reader to this figure.
    \item How natural the transition is between text and the figure, and whether the discussion around it is sufficient.  
\end{itemize}

Scoring:
\begin{itemize}
    \item \textbf{9-10 – Seamless Integration}: The text explicitly references the figure (e.g., “As shown in Figure 3…”), and it meaningfully discusses key elements of the figure, so the reader is guided to look at and interpret it.
    \item \textbf{7-8 – Good Integration}: There is a clear reference to the figure, but the transition is slightly abrupt, or the discussion is a bit brief or incomplete.
    \item \textbf{5-6 – Weak Link}: The figure is mentioned (e.g., only “(see Figure 2)” at the end of a sentence), but the surrounding text does not really explain or interpret it.
    \item \textbf{3-4 – Disconnected}: No explicit reference; the figure appears between paragraphs or at the side, and the relation has to be guessed by the reader.
    \item \textbf{1-2 – Disruptive}: The placement of the figure breaks the reading flow, feels random, or even contradicts the narrative sequence.
\end{itemize}

\textbf{3. Visual Information Value}

\textit{Does the figure provide visual information that text alone cannot effectively convey?}

Focus:
\begin{itemize}
    \item Whether the figure offers \textbf{unique visual value} that would be difficult, inefficient, or impossible to communicate through text alone.
    \item Consider how much understanding would be lost if the figure were removed? Would readers struggle to grasp the information from the text alone?
\end{itemize}

Scoring:
\begin{itemize}
    \item \textbf{9-10 – Visually Essential}: The figure conveys information that is extremely difficult or impractical to describe in text. Without the visual, readers would struggle to grasp the information even with lengthy description.
    \item \textbf{7-8 – High Visual Value}: The figure significantly enhances understanding by providing visual information that would require extensive text to approximate. Text alone would be notably less efficient or clear.
    \item \textbf{5-6 – Moderate Visual Value}: The figure provides some visual convenience, but most information could be reasonably conveyed in a few sentences. The visual format is helpful but not particularly necessary.
    \item \textbf{3-4 – Low Visual Value}: The figure adds minimal visual information beyond what text easily provides. The visual medium is barely justified.
    \item \textbf{1-2 – No Visual Necessity}: The figure provides no meaningful visual information that text cannot easily convey. Using a visual medium here is unnecessary regardless of content relevance.
\end{itemize}

\textbf{Instructions}
\begin{itemize}
    \item \textbf{Step 1}: Review the provided \textbf{Figure} and \textbf{Context}.
    \item \textbf{Step 2}: Evaluate the \textbf{Integration between the Figure and the Context} across the three dimensions: \textbf{Contextual Relevance}, \textbf{Narrative Coherence}, and \textbf{Visual Information Value}.
    \item \textbf{Step 3}: For each dimension, provide a \textbf{detailed justification} that examines concrete evidence from the figure and context, \textbf{then} explains the reasoning leading to the score.
    \item \textbf{Step 4}: Based on the justification, assign a \textbf{1–10 score} for each dimension according to the detailed scoring criteria.
    \item \textbf{Step 5}: If any dimension scores \textbf{6 or lower}, use the \textbf{suggestion} field to propose concrete improvements. If \textbf{all} dimensions score \textbf{7 or higher}, set the \textbf{suggestion} field as an empty string "".
\end{itemize}

\textbf{Output Format}
\vspace{0.5em}

Respond with a JSON object. Do not add markdown code blocks (like ```json) around the output.
\begin{lstlisting}[style=jsonstyle]
{
  "justification": {
    "Scoring Dimension": "specific reasons for the score...",
    ...
  },
  "scores": {
    "Scoring Dimension": 8,
    ...
  },
  "suggestion": "..."
}
\end{lstlisting}

\vspace{0.5em}
\textbf{USER PROMPT}
\vspace{0.5em}

Please evaluate the integration between the figure and the context based on the provided rubric.

\begin{verbatim}
<context>
{context}
</context>
\end{verbatim}
Remember to output ONLY the JSON object.

\textit{The user input also includes the figure itself.
}

\end{promptbox}

\begin{promptbox}{Prompt for Chart--Source Consistency}
\textbf{SYSTEM PROMPT}
\vspace{0.5em}

You are an expert Data Provenance \& Chart Verification Auditor. Your task is to evaluate \textbf{Chart-Source Consistency} in a research report using only the provided inputs.

You may be provided with:
\begin{enumerate}
    \item \textbf{Chart}: The chart itself.
    \item \textbf{Caption}: The full text associated with the chart.
    \item \textbf{References}: A JSON list of cited source snippets associated with the chart. Treat this list as the only permissible evidence; do not assume access to external links or any information not contained in the provided snippets.
\end{enumerate}

\textbf{Goal}
\vspace{0.5em}

Decide whether the chart is \textbf{consistent} with the provided sources.

\vspace{0.5em}
\textbf{Out Of Scope}
\begin{itemize}
    \item Pure visual aesthetics issues (layout, fonts, colors) unless they prevent verification.
    \item Real-world credibility of sources; only whether the provided snippets support the chart as presented.
    \item Missing or insufficient evidence. Only focus on \textbf{contradictions} between the chart and the provided sources.
\end{itemize}

\textbf{Definition Of An Issue}
\vspace{0.5em}

An issue exists when at least one claim from the chart is \textbf{contradicted} by References under the same (or materially equivalent) claim context.

\vspace{0.5em}
\textbf{Counting Rules}
\begin{itemize}
    \item Each distinct underlying problem counts as ONE issue.
    \item Multiple symptoms from the same root cause count as ONE issue.
    \item If there are multiple unrelated contradictions, count them as separate issues.
\end{itemize}

\textbf{Instructions}
\begin{itemize}
    \item \textbf{Step 1}: Read the \textbf{Chart} and \textbf{Caption} together to extract the chart’s claim(s).
    \item \textbf{Step 2}: Use the provided \textbf{References} as the only evidence to verify those claim(s).
    \item \textbf{Step 3}: Report every distinct issue you find, each with: 
    \begin{itemize}
        \item the chart claim,
        \item a short description,
        \item direct quote(s) from References with ref\_idx.
    \end{itemize}
    \item \textbf{Step 4}: Count the total number of distinct issues.
\end{itemize}

\textbf{Output Format}
\vspace{0.5em}

Respond with a JSON object. Do not add markdown code blocks (like ```json) around the output.
\begin{lstlisting}[style=jsonstyle]
{
  "total_issues": 5,
  "issues": [
    {
      "id": 1,
      "chart_claim": "the specific claim from the chart being checked",
      "description": "brief summary of the issue",
      "evidence": [
        {
          "ref_idx": "1",
          "quote": "direct quote from the corresponding reference snippet"
        },
        ...
      ]
    }
    ...
  ]
}
\end{lstlisting}

\vspace{0.5em}
\textbf{USER PROMPT}
\vspace{0.5em}

Please evaluate the Chart-Source Consistency based on the provided rubric.

\begin{verbatim}
<caption>
{caption}
</caption>

<references>
{json.dumps(references_payload, ensure_ascii=False)}
</references>
\end{verbatim}

Remember to output ONLY the JSON object. Do not include any other text or markdown formatting.

\textit{The user input also includes the chart itself.
}
\end{promptbox}
\subsection{Prompts for Report Generation}
\label{app:report_generation_prompts}

\begin{promptbox}{Prompt for Manus-1.6, Gemini-3-pro Deep Research, and Perplexity Deep Research}
You are responsible for generating well-structured, thoroughly argued, and richly illustrated multimodal deep research reports that meet the following requirements:

\begin{enumerate}
    \item The report must truly achieve rich illustration with text and images. At key argumentation points, you should \textbf{proactively and adequately insert visual elements} and embed them into the report. Visualization forms include but are not limited to charts generated using tools like Matplotlib, such as bar charts, line charts, heatmaps, radar charts, and related formats, as well as external images obtained through online searches, such as flowcharts, diagrams, architecture diagrams, and landscape images
    \item All \textbf{images and charts} in the report must have clear and uniformly formatted numbering and titles, and must be explicitly referenced and interpreted in the main text, for example, ``as shown in Figure 1''
    \item All \textbf{facts, data, viewpoints, images, and other content} obtained through online searches must be marked in the main text with \textbf{standardized reference citations}, for example, ``\ldots[1]''; charts generated using tools must also be annotated according to this requirement with their underlying data sources
    \item At the end of the report, complete reference entries corresponding to the numbering in the main text must be listed according to a unified academic format, including accessible \textbf{URLs}
    \item The report must be written in \textbf{Markdown} and follow standard Markdown syntax
\end{enumerate}
\end{promptbox}

\begin{promptbox}{Prompt for Grok-4.1-Thinking DeepSearch}
You are responsible for generating well-structured, thoroughly argued, and richly illustrated multimodal deep research reports that meet the following requirements:

\begin{enumerate}
    \item The report must truly achieve rich illustration with text and images. At key argumentation points, you should \textbf{proactively and adequately insert visual elements} and embed them into the report. Visualization forms include but are not limited to charts generated using tools like Chart.js, such as bar charts, line charts, heatmaps, radar charts, and related formats, as well as external images obtained through online searches, such as flowcharts, diagrams, architecture diagrams, and landscape images
    \item All \textbf{images and charts} in the report must have clear and uniformly formatted numbering and titles, and must be explicitly referenced and interpreted in the main text, for example, ``as shown in Figure 1''
    \item All \textbf{facts, data, viewpoints, images, and other content} obtained through online searches must be marked in the main text with \textbf{standardized reference citations}, for example, ``\ldots[1]''; charts generated using tools must also be annotated according to this requirement with their underlying data sources
    \item At the end of the report, complete reference entries corresponding to the numbering in the main text must be listed according to a unified academic format, including accessible \textbf{URLs}
    \item The report must be written in \textbf{Markdown} and follow standard Markdown syntax
\end{enumerate}
\end{promptbox}

\begin{promptbox}{Prompt for Claude-4.5-Sonnet w/Search and Genspark Deep Research}
You are responsible for generating well-structured, thoroughly argued, and richly illustrated multimodal deep research reports that meet the following requirements:

\begin{enumerate}
    \item The report must truly achieve rich illustration with text and images. At key argumentation points, you should \textbf{proactively and adequately insert visual elements} and embed them into the report. Visualization forms include but are not limited to charts generated using tools like Chart.js, such as line charts, heatmaps, radar charts, and related formats, as well as external images obtained through online searches, such as diagrams, architecture diagrams, and landscape images
    \item All \textbf{images and charts} in the report must have clear and uniformly formatted numbering and titles, and must be explicitly referenced and interpreted in the main text, for example, ``as shown in Figure 1''
    \item All \textbf{facts, data, viewpoints, images, and other content} obtained through online searches must be marked in the main text with \textbf{standardized reference citations}, for example, ``\ldots[1]''; charts generated using tools must also be annotated according to this requirement with their underlying data sources
    \item At the end of the report, complete reference entries corresponding to the numbering in the main text must be listed according to a unified academic format, including accessible \textbf{URLs}
    \item The report must be written in \textbf{HTML} and follow standard HTML syntax
\end{enumerate}
\end{promptbox}

\end{document}